\documentclass[10pt,twocolumn,letterpaper]{article}

\usepackage[pagenumbers]{cvpr}

\newcommand{\mt}[1]{#1}
\usepackage{multirow}
\usepackage{makecell}
\usepackage{etoc}
\usepackage{fontawesome5}
\newcommand{\egoicon}{{\footnotesize\faGlasses}\,}
\newcommand{\exoicon}{{\footnotesize\faVideo}\,}

\usepackage[accsupp]{axessibility}

\definecolor{cvprblue}{rgb}{0.21,0.49,0.74}
\usepackage[pagebackref,breaklinks,colorlinks,allcolors=cvprblue]{hyperref}

\usepackage{orcidlink}

\def\confName{CVPR}
\def\confYear{2027}

\title{Do Egocentric Video-Language Models Capture \\Both Hand- and Object-Centric Cues?}

\author{
Masatoshi Tateno$^{1}$\orcidlink{0009-0001-2301-649X} \quad
Alexandros Stergiou$^{2}$\orcidlink{0000-0003-4706-4231} \quad
Risa Shinoda$^{1}$\orcidlink{0009-0006-3965-7933} \quad
Yoichi Sato$^{1}$\textsuperscript{\dag}\orcidlink{0000-0003-0097-4537} \quad
Dima Damen$^{3}$\textsuperscript{\dag}\orcidlink{0000-0001-8804-6238}
\\[2pt]
$^{1}$The University of Tokyo \qquad
$^{2}$University of Twente \qquad
$^{3}$University of Bristol
}

\begin{document}
\maketitle

\iftoggle{cvprfinal}{%
  \begingroup
  \renewcommand\thefootnote{\dag}%
  \footnotetext{Equal supervision.}%
  \endgroup
}{}

\begin{abstract}

Recognizing hand-object interactions (HOIs) requires capturing both hand dynamics and object transformations, which provide complementary information for distinguishing visually similar manipulations. However, Video-Language Models (VLMs) trained solely on egocentric videos tend to over-rely on hand appearance, struggling to distinguish interactions with similar visual appearances but different dynamics (e.g., taking vs. kneading dough).
To better utilize complementary hand- and object-centric cues, we propose a learning paradigm with \mt{(i) hand-object masked training to encode both hand- and object-centric cues and learn their relationships through cross-cue reconstruction, and (ii) an HOI-dynamics-aware decoder that grounds each cue through auxiliary location and semantic predictions.}
To assess how effectively models leverage hand- and object-centric cues for action recognition, we introduce \textbf{C}ue-\textbf{I}solated \textbf{HOI} (\textbf{CI-HOI}), which independently evaluates recognition performance based on each cue.
To enable this evaluation, we curate the DEHOI testbed using video inpainting to selectively remove either hands or manipulated objects, allowing us to evaluate action recognition with each cue isolated.
DEHOI reveals that existing VLMs trained only on egocentric video rely more on hand-centric cues than VLMs trained on more non-egocentric video data.
We show that our training paradigm mitigates this hand-centric bias, outperforming VLMs trained on the same egocentric data across DEHOI, standard action recognition, and object state recognition benchmarks.

\end{abstract}

\section{Introduction}
\label{sec:intro}

Hand-Object Interaction (HOI) recognition is essential for understanding human actions and their effects on the environment.
In particular, hand dynamics and object transformations provide complementary cues, both of which are important for accurately and robustly distinguishing fine-grained manipulations.

\begin{figure}[t]
    \centering
    \includegraphics[width=1\linewidth]{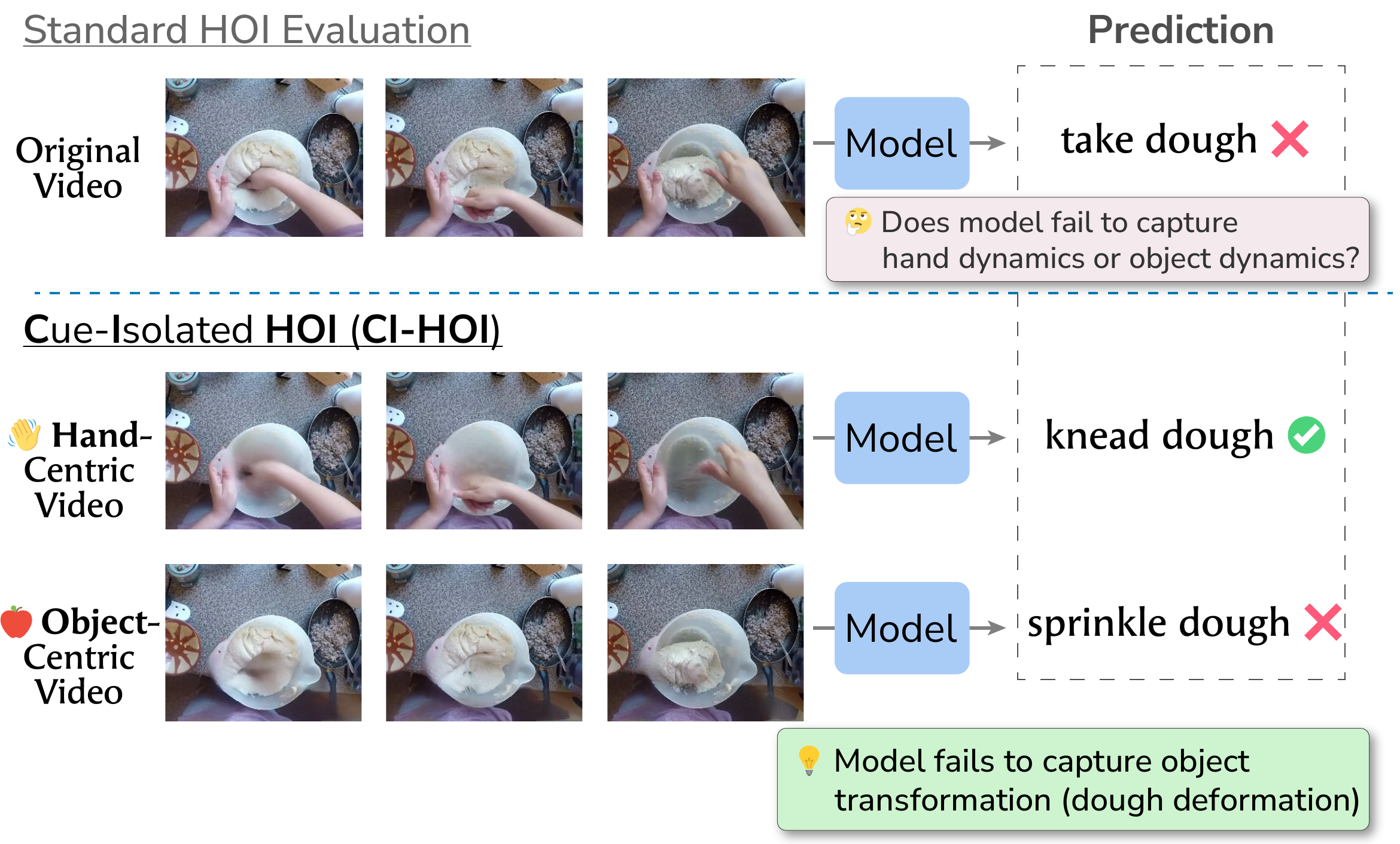}
    \caption{\textbf{C}ue-\textbf{I}solated \textbf{HOI}. Our new CI-HOI evaluation assesses models’ understanding of hand dynamics and object transformations through verb prediction from cue-isolated videos. In this example, recognition failure can be attributed to the incorrect recognition of \emph{dough} transformation.}
    \label{fig:dehoi}
\end{figure}

However, existing egocentric Video-Language Models (VLMs) often confuse visually similar actions~\cite{xu2024egocentric} that share hand postures and object categories but differ in their underlying interactions (e.g., in~\cref{fig:dehoi} top, the model recognizes \textit{knead} as \textit{take}). 
This suggests that current models tend to over-rely on correlations between the appearances of hands and manipulated objects, without sufficiently capturing the dynamics of hand manipulations and object transformations that distinguish these actions.

Prior studies have sought to improve sensitivity to subtle action differences through language-side supervision, such as using similar yet distinct verbs or captions as negatives~\cite{zhao2023learning,pei2025modeling,xu2024egocentric}.
However, how to better capture hand- and object-centric visual cues within the visual encoder, and how to systematically evaluate the encoder's behavior in capturing these cues, remain underexplored.

To better capture these visual cues, we propose a framework that explicitly learns both hand- and object-centric cues and grounds their dynamics in interaction semantics.
First, we introduce hand-object masked training, which reconstructs masked hand or object patches from the remaining visual information to encourage learning cross-cue relationships.
Second, our HOI-dynamics-aware (HDA) decoder grounds each cue in spatial and semantic information through a multi-task objective. The decoder jointly predicts hand-object locations and object/action categories while aligning video representations with the overall interaction narration. Together, these components complement each other by learning cross-cue relationships while explicitly grounding each cue, discouraging reliance on superficial shortcuts.

To systematically investigate how visual encoders respond to hand- and object-centric cues, we introduce \textbf{C}ue-\textbf{I}solated \textbf{HOI} (\textbf{CI-HOI}), a new evaluation paradigm that probes encoder sensitivity to each cue through selective visual ablation.
We curate the \textbf{DEHOI} testbed by selectively inpainting either hands or manipulated objects, enabling controlled evaluation of action recognition under isolated visual cues (see~\cref{fig:dehoi}).

Evaluation on DEHOI reveals that VLMs trained exclusively on egocentric videos exhibit a stronger reliance on hand-centric cues than those trained with more non-egocentric video data.
Our training paradigm mitigates this hand-centric bias by explicitly supervising both cues, enabling more balanced recognition of hand- and object-specific dynamics compared to existing models.
These improvements also hold for original videos, yielding consistent gains in standard action recognition.
We further observe improvements in object-state classification and state-change identification, suggesting stronger modeling of object transformations.

In summary, we:

\begin{enumerate}

    \item Propose a hand-object-aware learning paradigm with (i) hand-object masked training and (ii) an HOI-dynamics-aware decoder. Together, they effectively and explicitly capture hand- and object-centric cues.

    \item Introduce \textbf{CI-HOI} and the \textbf{DEHOI} testbed to systematically probe visual encoder sensitivity to hand- and object-centric cues through selective visual ablation.

    \item Reveal the hand-centric bias of VLMs trained solely on egocentric videos and demonstrate that our approach mitigates this bias while improving action and object-state recognition.

\end{enumerate}

\section{Related Work}
\label{sec:related_work}

\subsection{Video-Language Modeling for HOI Understanding}
\mt{
Large-scale egocentric video-text datasets such as Ego4D~\cite{grauman2022ego4d} have enabled video-language models including EgoVLP~\cite{lin2022egocentric} and EgoVLPv2~\cite{pramanick2023egovlpv2}.
Recent contrastive vision-language learning approaches have used LLM-enriched captions and harder positive and negative samples~\cite{zhao2023learning, pei2025modeling, xu2024egocentric, mandikal2026spoc}. Helping Hands~\cite{zhang2023helping} incorporates hand/object locations and object categories to improve spatial grounding.
However, these approaches place less emphasis on explicitly learning hand- and object-centric dynamics as complementary cues in visual representations.
}

\mt{
\subsection{Disentangling Visual Cues}
Prior work has decomposed visual cues for action recognition along two main axes.
First, studies on human dynamics and static scene context exposed and mitigated models’ reliance on background shortcuts~\cite{choi2019can, wang2018pulling, bae2024devias, bae2025mash, kowal2022deeper, qian2022static}.
Second, compositional action recognition (CAR) decomposed actions into verbs and nouns and generalized to unseen compositions~\cite{materzynska2020something, luo2022disentangled, li2024c2c}.
However, these approaches either suppress static context or decompose the label space. They do not reveal how models use hand- and object-centric visual evidence. In contrast, we decompose HOI at the visual level into \textit{hand and object dynamics}, two complementary cues underlying manipulation.
Our training encourages the model to capture both cues and their relationships, while CI-HOI probes how the model responds to each cue through selective visual ablation.
}

\subsection{Hand-Object Interaction Understanding Benchmarks}

Egocentric video benchmarks evaluate HOI understanding through action recognition~\cite{goyal2017something, li2018eye, damen2018scaling, damen2022rescaling, perrett2025hd}, typically representing actions as verb-noun pairs. EgoMCQ~\cite{lin2022egocentric} and EPIC-KITCHENS 100~\cite{damen2022rescaling} assess high-level semantics through video retrieval and grounding.
Another line focuses on changes in object states. Ego4D~\cite{grauman2022ego4d} is used for temporal, spatial, and semantic understanding, and ChangeIt~\cite{souvcek2022look} and HowToChange~\cite{xue2024learning} for state transitions, with extensions to multi-state~\cite{tateno2025learning}, multi-view~\cite{hong2021transformation,saini2023chop}, and multi-object~\cite{zameni2025moscato} settings. HandyVQA~\cite{tateno2026handyvqa} further covers both actions and object states through video QA on fine-grained dynamics.
Despite these advances, existing benchmarks evaluate hand and object cues jointly, as they co-occur in the same video.
Our CI-HOI instead isolates these cues at the pixel level to probe how visual encoders respond to each cue.

\section{Hand-Object-Aware Robust Representation Learning}
\mt{Our approach leverages hand-object complementarity to learn robust cue-specific representations from \emph{unedited} egocentric videos.
Our approach consists of:
(i) \emph{Masked Training} (\cref{subsec:hand-object_masked_training}), which masks hand or object regions and reconstructs them from the remaining cues to capture individual cues and their relationships; and
(ii) an \emph{HOI-Dynamics-Aware (HDA) Decoder} (\cref{subsec:hda_decoder}), which promotes cue-specific semantic reasoning by learning hand- and object-centric embeddings aligned with their locations and semantics.
We describe each component next.}

\begin{figure*}[t]
    \centering
    \includegraphics[width=1\linewidth]{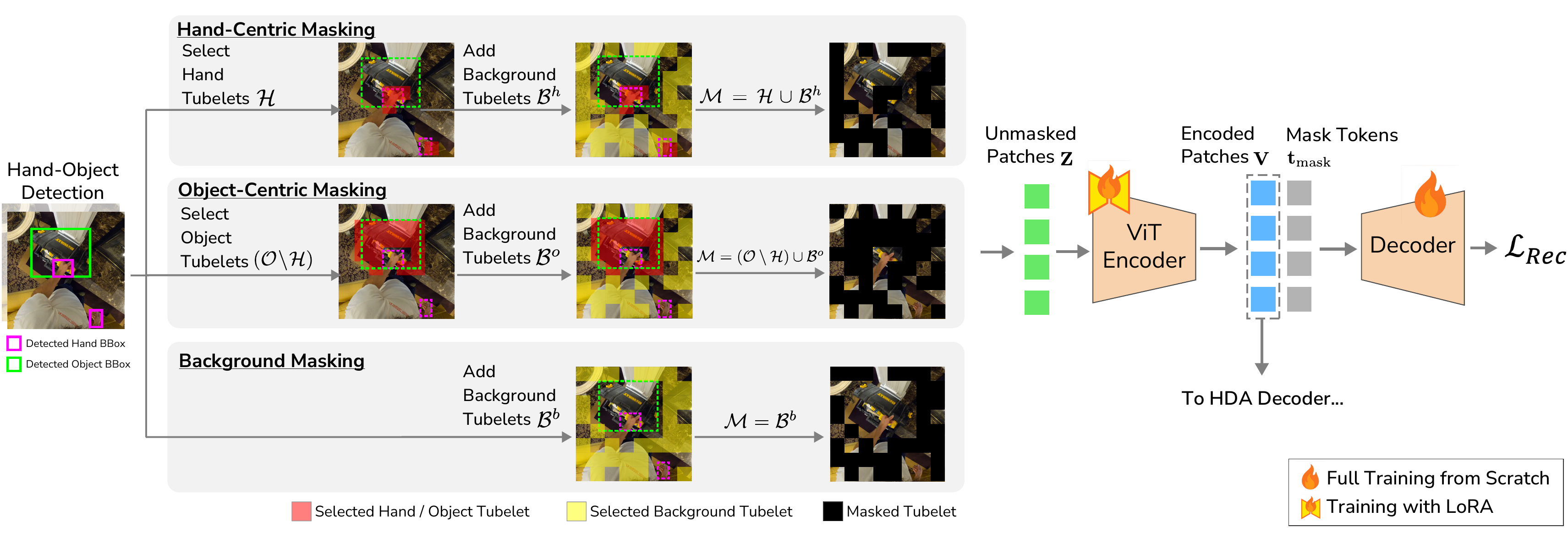}
    \vspace{-10pt}
    \caption{\mt{Hand-object masked training. Input visual patches are masked and reconstructed. The masking strategy enforces hand- and object-centric or background regions as reconstruction context/targets (left). $\mathcal{H}$ and $\mathcal{O}$ denote sets of hand and object tubelet indices, respectively $\mathcal{R}$ denotes a set of randomly selected tubelet indices, and $\mathcal{M}$ denotes the final set of tubelets to mask.}}
    \label{fig:method_masking}
\end{figure*}

\subsection{Hand-Object Masked Training}
\label{subsec:hand-object_masked_training}
\subsubsection{Preliminary: Masked Autoencoder.}
Video Masked Autoencoders (VideoMAE) are self-supervised pretext tasks that learn video representations by reconstructing masked spatiotemporal patches~\cite{tong2022videomae}.
\mt{We overview our masked training in~\cref{fig:method_masking}.}
Given a video $\mathbf{x}\in\mathbb{R}^{T\times H\times W\times C}$, we partition it into $N$ non-overlapping space-time patches $\{\mathbf{p}_i\}_{i=1}^{N}$ of size $(p_t,p_h,p_w)$. 
Each patch is vectorized as $\mathbf{P}_i=\mathrm{vec}(\mathbf{p}_i)\in\mathbb{R}^{d_p}$, where $d_p=p_t p_h p_w C$, and projected via a learnable projection (e.g., linear layer or convolution) to obtain token embeddings $\mathbf{z}_i=\texttt{proj}(\mathbf{P}_i)\in\mathbb{R}^{D}$. Tokens are dropped given
a set of indixes $\mathcal{M}\subset\{1,\dots,N\}$. They are randomly sampled with mask ratio $r=|\mathcal{M}|/N$ resulting in $\mathcal{V}=\{1,\dots,N\}\setminus\mathcal{M}$ visible tokens.
The encoder $f$ embeds only the non-masked subset of tokens $\mathbf{Z}=\{\mathbf{z}_i\}\forall i \in \mathcal{V}$ with corresponding position encodings $\texttt{pos}_{f}$:
\[
\mathbf{V}=f(\mathbf{Z}+\texttt{pos}_{f})\in\mathbb{R}^{|\mathcal{V}|\times D_e},
\]
Using embeddings $\mathbf{V}$ as context alongside learnable mask tokens $\mathbf{t}_{\mathrm{mask}}$ at masked positions, a decoder $g$ reconstructs the masked patches. The encoder and decoder are trained end-to-end with a mean-square error loss between the reconstructed masked tokens at the embedding level $\mathbf{z}_i$. The decoder uses $\mathbf{Y} = \mathbf{V} \cup {\mathbf{t}_{\mathrm{mask}}}$ that combine encoded visible latents $\mathbf{S}$ and learnable mask tokens $\mathbf{t}_{\mathrm{mask}}$:
\[
\hat{\mathbf{Z}}=g(\mathbf{Y}+\texttt{pos}_g)\in\mathbb{R}^{N\times D}
\]
The objective minimizes the reconstruction error on masked tokens $\mathbf{\hat{\mathbf{z}}_i} \in \hat{\mathbf{Z}} $:
\[
\mathcal{L}_{\mathrm{Rec}}=\frac{1}{|\mathcal{M}|}\sum_{i\in\mathcal{M}}\|\mathbf{z}_i-\hat{\mathbf{z}}_i\|_2^2
\]
which trains the encoder to learn representations predicting missing spatiotemporal tokens from the visible context.

\begin{figure*}[t]
    \centering
    \includegraphics[width=1\linewidth]{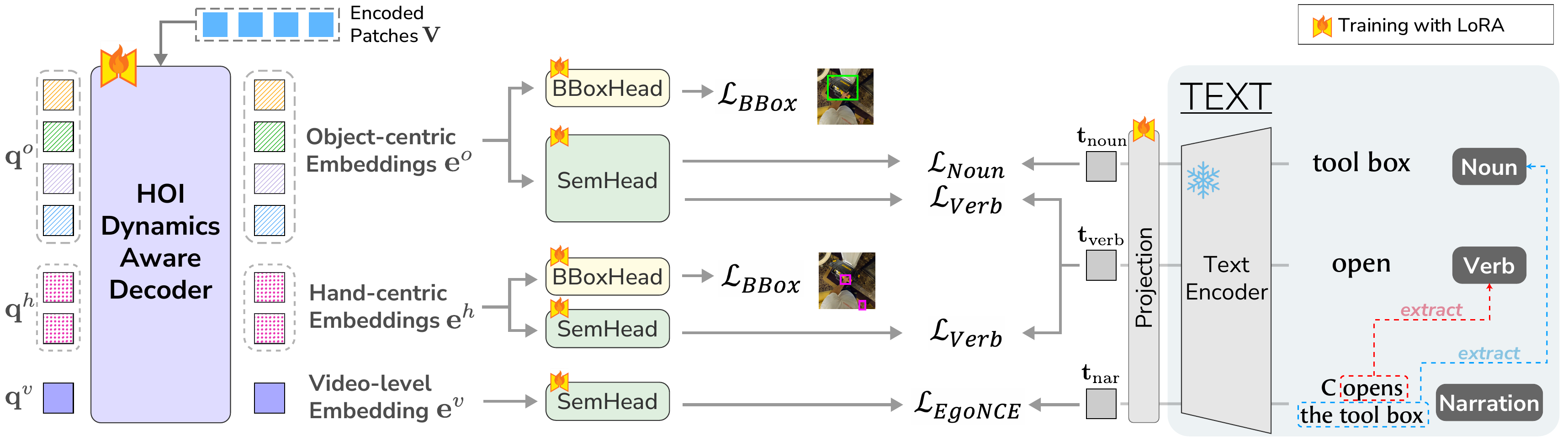}
    \vspace{-10pt}
    \caption{Training HOI-dynamics-aware (HDA) decoder. Hand- and object-centric queries cross-attend to encoded visual patches to produce cue-specific embeddings, supervised by hand/object locations and semantics. The video-level embedding interacts with these embeddings and visual patches to encode hand-object dynamics, and is aligned with narration summarizing the overall interaction.}
    \label{fig:method_hda}
\end{figure*}

\subsubsection{Hand-Object Masking.}
\mt{To encode individual hand- and object-specific cues and their relationships, we design a masking strategy for $\mathcal{M}$ that selectively drops hand, manipulated-object, or background regions\mt{, as illustrated on the left of~\cref{fig:method_masking}}.
For this, we use an off-the-shelf HOI detector~\cite{shan2020understanding} on the video to obtain bounding boxes for hands and manipulated objects.}

\noindent
\textbf{Tubelet sets.}
Following the tubelet tokenization described above, each token corresponds to a
spatiotemporal patch $\mathbf{p}_{(t,s)}$ of size $(p_t,p_h,p_w)$, where
$s$ indexes the spatial location and $t$ indexes the temporal segment.
We group tokens that share the same spatial location across time into a spatial tubelet $T_s$, i.e., $T_s = \{\mathbf{p}_{(t,s)} \mid t = 1,\dots,T/p_t\}$.
Let $R_s^{(t)}$ denote the spatial region of tubelet $T_s$ at frame $t$, and
let $H_t$ and $O_t$ be the hand and object bounding boxes at frame $t$.
We assign $T_s$ as a hand-related tubelet if the hand-tubelet overlap exceeds $50\%$ in at least one frame, $\max_{t} \frac{|H_t \cap R_s^{(t)}|}{|R_s^{(t)}|} > 0.5$, and as object-related if $\max_{t} \frac{|O_t \cap R_s^{(t)}|}{|R_s^{(t)}|} > 0.5$.
We denote the resulting index sets as $\mathcal{H}$ and $\mathcal{O}$, and define background tubelets as $\mathcal{B} = \{1,\dots,N_s\}\setminus(\mathcal{H}\cup\mathcal{O})$, where $N_s = \frac{H}{p_h}\frac{W}{p_w}$ is the number of spatial tubelets.
Note that a tubelet may satisfy both conditions, i.e., $s \in \mathcal{H} \cap \mathcal{O}$.

\noindent
\textbf{Hand-centric masking.}
We define the masked set as 
$\mathcal{M} = \mathcal{H} \cup \mathcal{B}^h$, 
where $\mathcal{B}^h \subseteq \mathcal{B}$ 
is randomly sampled until $|\mathcal{M}|=\tau N_s$. 
Since all hand tubelets are already masked, additional masking is performed only over background tubelets, ensuring that object cues remain visible.

\noindent
\textbf{Object-centric masking.}
We define 
$\mathcal{M} = (\mathcal{O}\setminus\mathcal{H}) \cup \mathcal{B}^o$, 
where $\mathcal{B}^o \subseteq \mathcal{B}$ 
is randomly sampled until $|\mathcal{M}|=\tau N_s$. 
We exclude hand tubelets since detected hands typically appear in front of the manipulated objects, and additional masking is restricted to background tubelets to preserve hand cues.

\noindent
\textbf{Background masking.}
We define 
$\mathcal{M} = \mathcal{B}^b$, 
where $\mathcal{B}^b \subseteq \mathcal{B}$ 
is randomly sampled until $|\mathcal{M}|=\tau N_s$.

\noindent
The encoder processes only the unmasked tubelets $\{1,\dots,N_s\}\setminus\mathcal{M}$, whose outputs are fed to the HOI-dynamics-aware decoder described next.

\subsection{HOI-Dynamics-Aware Decoder}
\label{subsec:hda_decoder}

\mt{Masked training encourages the encoder to capture individual hand and object cues and their relationships. We further aim to explicitly associate each cue with its role in the interaction.
We introduce an HOI-dynamics-aware (HDA) decoder that grounds hand- and object-centric representations in their locations and semantics, promoting cue-specific reasoning about HOI dynamics.

As illustrated in~\cref{fig:method_hda}, HDA is a DETR-like Transformer decoder~\cite{carion2020end} with learnable hand, object, and video queries $[\mathbf{q}^h;\mathbf{q}^o;\mathbf{q}^v]$.
While structurally related to the object-aware decoder in Helping Hands~\cite{zhang2023helping}, HDA explicitly learns distinct hand- and object-centric representations and grounds both in interaction dynamics.
Specifically, we use two hand queries $\mathbf{q}^h \in \mathbb{R}^{2\times512}$, $K$ object queries $\mathbf{q}^o \in \mathbb{R}^{K\times512}$, and one video query $\mathbf{q}^v \in \mathbb{R}^{1\times512}$.
The queries cross-attend to encoded visual patches $\mathbf{V}$ and interact through self-attention to produce hand-centric $\mathbf{e}^h$, object-centric $\mathbf{e}^o$, and video-level $\mathbf{e}^v$ embeddings:
\begin{equation}[\mathbf{e}^h;\mathbf{e}^o;\mathbf{e}^v] = \mathrm{HDA}([\mathbf{q}^h;\mathbf{q}^o;\mathbf{q}^v],\mathbf{V}).
\end{equation}
We supervise the cue-specific embeddings with their locations and semantics, while aligning the video-level embedding with the narration describing the overall interaction.}

\noindent
\textbf{Spatial grounding.}
We train $\mathbf{e}^h$ and $\mathbf{e}^o$ to localize hands and manipulated objects at each frame.
Since these embeddings summarize the video over space and time, we concatenate them with learnable frame embeddings $\mathbf{e}_t$, yielding
$\widehat{\mathbf{e}}^h=[\mathbf{e}^h;\mathbf{e}_t]$ and
$\widehat{\mathbf{e}}^o=[\mathbf{e}^o;\mathbf{e}_t]$.
An MLP head $\mathcal{H}_{\mathrm{bbox}}$ regresses frame-wise hand and object boxes.
After Hungarian matching at each frame, we optimize
\begin{equation}
\mathcal{L}_{\mathrm{BBox}}
=
\lambda_{\ell_1}\lVert\hat{\mathbf{b}}-\mathbf{b}\rVert_1
+
\lambda_{\mathrm{giou}}\mathcal{L}_{\mathrm{giou}}(\hat{\mathbf{b}},\mathbf{b}).
\end{equation}

\noindent
\textbf{Semantic grounding.}
A semantic head $\mathcal{H}_{\mathrm{sem}}$ projects the cue-specific embeddings to
$\mathbf{s}^h=\mathcal{H}_{\mathrm{sem}}(\mathbf{e}^h)$ and
$\mathbf{s}^o=\mathcal{H}_{\mathrm{sem}}(\mathbf{e}^o)$,
where $\mathbf{s}^h,\mathbf{s}^o\in\mathbb{R}^{216}$.
We align both hand- and object-centric embeddings with verb semantics to associate each cue with interaction dynamics, while additionally aligning object-centric embeddings with noun semantics.

For noun supervision, we use Hungarian matching based on cosine similarity to align object-query embeddings with ground-truth noun embeddings, and apply an NCE loss to the matched pairs:
\begin{equation}
\mathcal{L}_{\mathrm{Noun}}
=
\frac{1}{K}\sum_{j=1}^{K}
\mathrm{NCE}(\mathbf{s}^{o}_j,\mathbf{t}_{\mathrm{noun}}^{(\sigma(j))}),
\end{equation}
where $\sigma$ denotes the Hungarian matching.

To explicitly associate both hand and object cues with interaction dynamics, we apply verb-level NCE to both types of embeddings:
\begin{equation}
\mathcal{L}_{\mathrm{Verb}}
=
\frac{1}{N_h}\sum_{i=1}^{N_h}
\mathrm{NCE}(\mathbf{s}^{h}_i,\mathbf{t}_{\mathrm{verb}})
+
\frac{1}{K}\sum_{j=1}^{K}
\mathrm{NCE}(\mathbf{s}^{o}_j,\mathbf{t}_{\mathrm{verb}}),
\end{equation}
where $N_h$ is the number of hand queries.
When multiple verbs are valid, all corresponding verb embeddings are treated as positives.
Together, the spatial and semantic objectives encourage each cue-specific embedding to encode both where the interaction occurs and what dynamics it represents.

\subsubsection{Video-Text Alignment.}
Finally, the video-level embedding integrates information from the cue-specific embeddings and visual patches through decoder self- and cross-attention.
We align this representation with narrations using EgoNCE~\cite{lin2022egocentric,zhao2023learning}, which captures the semantics of the overall interaction.
Following its standard formulation, samples sharing at least one noun and verb are treated as additional positives, while narrations from different temporal segments of the same video provide hard negatives.
We denote the resulting symmetric video-text objective as $\mathcal{L}_{\mathrm{EgoNCE}}$.

\subsection{Aggregated Objective}
The overall training objective is a weighted sum of video-text alignment, semantic supervision, spatial localization, and reconstruction losses:
\begin{equation}
\begin{split}
\mathcal{L}_{\text{Total}}
= \;&
\lambda_{\text{vt}} \, (\mathcal{L}_{\text{EgoNCE}}^{v \rightarrow t} + \mathcal{L}_{\text{EgoNCE}}^{t \rightarrow v})
+
\lambda_{\text{Noun}} \, \mathcal{L}_{\text{Noun}}
+
\lambda_{\text{Verb}} \, \mathcal{L}_{\text{Verb}}
\\
&+
\mathcal{L}_{\text{BBox}}
+
\mathcal{L}_{\text{Rec}},
\end{split}
\end{equation}
where $\lambda_{x}$ are weighting coefficients.

\subsection{Inference}
During inference, we do not apply any masking to visual patches and encode the entire video.
In the HDA decoder, the video-level embedding projected into the joint visual-textual space through the semantics head is used for downstream tasks, while the hand- and object-centric embeddings are not utilized.

\section{CI-HOI: Cue-Isolated HOI Recognition}
\label{sec:ci-hoi}

\noindent\textbf{CI-HOI}
evaluates hand- and object-centric reasoning in a disentangled manner.
Given HOI videos in which either the hands or the manipulated objects are removed, models are tasked with predicting the corresponding interaction verb from the remaining visual cues (see~\cref{fig:dehoi}).

\noindent\textbf{Motivation.}
By isolating hand- and object-centric visual cues, CI-HOI reveals potential shortcut behaviors and imbalanced reliance on specific visual cues.
Unlike conventional approaches that assess hand- or object-centric reasoning through semantic label prediction, CI-HOI isolates visual cues directly in pixel space, allowing us to explicitly examine which cues models rely on for HOI recognition.

\noindent\textbf{Scope.}
CI-HOI focuses on model sensitivity to hand- and target-object-centric visual cues, which primarily capture interaction dynamics, while leaving background cues outside the scope of evaluation.
During inference, the target object name is provided to focus the evaluation on distinguishing interaction verbs based on hand dynamics and object transformations.

\noindent\textbf{DEHOI Testbed.}
To enable CI-HOI, we curate the \textbf{DEHOI} testbed.
We source HOI videos from EPIC-KITCHENS-100 (EK-100)~\cite{damen2022rescaling} and obtain hand and target-object masks from VISOR~\cite{darkhalil2022epic}.
For frames without annotations, we propagate masks using SAM2~\cite{ravi2024sam} to obtain temporally consistent masks.
We then apply ProPainter~\cite{zhou2023propainter} to remove either the hands or target objects, producing object-centric and hand-centric videos, respectively.
DEHOI contains 2{,}652 cue-isolated videos, each paired with its target object name and interaction verb.

\begin{table}[t]
\centering
\resizebox{\linewidth}{!}{
\begin{tabular}{lcccccc}
\toprule
\multirow{2}{*}{Condition} & \multirow{2}{*}{Accuracy} & \multirow{2}{*}{$\Delta$}
& \multicolumn{3}{c}{Reported cue source (\%)} \\
\cmidrule(lr){4-6}
& & & Hand & Object & Could not tell \\
\midrule
Original & 73.6 {\small $\pm$5.8} & -- & -- & -- & -- \\
Hand-centric & 63.9 {\small $\pm$8.4} & -9.7 & 64.9 & 23.8 & 11.3 \\
Object-centric & 70.5 {\small $\pm$6.9} & -3.1 & 7.8 & 85.1 & 7.1 \\
\bottomrule
\end{tabular}
}
\vspace{-4pt}
\caption{\mt{Human study on DEHOI (96 clips, 16 verbs, 18 participants). Accuracy is averaged over participants (subscript: SD); $\Delta$ denotes the drop from original videos. In cue-isolated conditions, participants also reported the cue used to identify the action.}}
\label{tab:human_study}
\end{table}

\noindent
\textbf{Human study.}
\mt{We conduct a human study on DEHOI to assess interaction recognition from isolated cues and provide a human reference for CI-HOI.
We select the 10 most frequent verbs and six additional verbs for balance, sampling six clips per verb (96 clips total).
Each clip is presented in three conditions: unedited original, hand-centric with object inpainting, and object-centric with hand inpainting. No participant sees the same clip in multiple conditions.
As in model evaluation, the 18 participants receive the target object name and select a verb from 10 candidates (10\% chance).
In the two cue-isolated conditions, participants additionally report the cue they relied on, \ie, the visible hand, the visible object, or that they could not tell the action at all.
We collect responses from 18 participants.}

\mt{\cref{tab:human_study} summarizes the results.
Even for the original videos, human accuracy does not reach 100\%, indicating an inherent ambiguity in verb selection.
Human participants exhibit lower accuracy and greater uncertainty in the hand-centric condition than in the object-centric condition, with more frequent failures to identify the verb. Even in the hand-centric condition, participants report relying on object-related cues, such as surrounding tool motion.
These results suggest that humans rely more on object cues than on hand cues for action recognition.
\cref{sec:human_study_supp} details the interface and verb-wise results.}

\section{Experiments}
We overview model and training details in~\cref{sec:experiments::implementation} alongside benchmark datasets used in~\cref{sec:experiments::datasets}. We present DEHOI results in~\cref{sec:experiments::dehoi} and performance comparisons across object state benchmarks in~\cref{sec:experiments::osr}. 

\subsection{Implementation Details}
\label{sec:experiments::implementation}
\noindent \textbf{Architecture.}
We adopt LaViLa~\cite{zhao2023learning} pre-trained on EgoClip~\cite{lin2022egocentric} as our video encoder and text encoder.
For hand-object masked training, we employ a lightweight Transformer decoder $g(\cdot)$ with one layer and four attention heads.
For HDA, we build upon the object-aware decoder of~\cite{zhang2023helping}, initializing it from their released checkpoint.
We fine-tune both the video encoder $f(\cdot)$ and the $\text{HDA}(\cdot,\cdot)$ decoder with Low-Rank Adapters (LoRA)~\cite{hu2022lora}, using a rank of 8 and a scaling factor $\alpha$ of 4; the layers to which LoRA is applied are detailed in~\cref{subsec:lora_supp}.

\noindent \textbf{Training.}
For each clip, we uniformly sample $4 \times 244 \times 244$ RGB frames.
We train the backbone and the HDA decoder jointly for two epochs with a batch size of 256.
We use AdamW~\cite{loshchilov2017decoupled} with a learning rate of $1\mathrm{e}{-5}$.
In masked training, the masking ratio $\tau$ is set to 0.5.
For each sample, the masking strategy is selected probabilistically: with 80\% probability, we apply a hand- or object-centric masking strategy, and with 20\% probability, a background masking strategy.
Within this strategy, hand and object regions are masked with probabilities 0.1 and 0.9. Loss weights are set to $\lambda_{\text{vt}}=0.2$, $\lambda_{\text{Noun}}=0.5$, $\lambda_{\text{Verb}}=0.3$.

\noindent \textbf{Inference.}
We adjust the number of input frames to follow the original protocol of each benchmark: 16 frames for DEHOI, 1 frame for static state classification, and 2 frames for state change identification in STATUS Bench.
For benchmarks with 1- or 2-frame inputs, we replicate each frame 4 and 2 times, respectively, to form a temporal dimension matching the input format of the pretrained model.

\subsection{Datasets and Metrics}
\label{sec:experiments::datasets}
For training, we use only EgoClip dataset, an egocentric video-text dataset built on Ego4D~\cite{grauman2022ego4d} containing 3.8M video-text pairs, following prior work~\cite{lin2022egocentric, zhao2023learning, zhang2023helping} to ensure a fair comparison.
For evaluation, we use \textbf{DEHOI} to assess hand-object reasoning under cue-isolated settings. 
Our analysis reveals that models trained only on egocentric videos tend to rely more on hand cues, underutilizing object information.
To further validate this observation, we evaluate object state (change) recognition on \textbf{STATUS Bench}~\cite{ukai2025status},
showing that our training paradigm more effectively captures object-centric cues than existing models.

\noindent
\textbf{STATUS Bench~\cite{ukai2025status}} contains 404 image pairs sourced from Ego4D~\cite{grauman2022ego4d} capturing object states before and after a change.
Each image is annotated with an object-state description, while image pairs are associated with ground-truth state-change descriptions and semantically similar four distractors.
The benchmark defines three tasks: (i) \emph{Object State Identification}, retrieving the correct state description for each image; (ii) \emph{Image Retrieval}, retrieving the correct image for a given state description; and (iii) \emph{State Change Identification}, selecting the correct state-change description from candidates given an image pair. Performance is measured using accuracy.

\subsection{Evaluation Protocol}

\noindent\mt{\textbf{Baselines.}
We compare with representative egocentric video models.
LaViLa~\cite{zhao2023learning} and Helping Hands~\cite{zhang2023helping} use the same backbone and training corpus as ours, with Helping Hands additionally using the same box supervision.
EgoVideo~\cite{pei2025modeling} represents large-scale dual-encoder models scaled to larger and more diverse corpora, while VideoChat2~\cite{li2024mvbench} and LLaVA-Video-7B~\cite{zhang2024llava} represent LLM-integrated models trained on large-scale video corpora including Ego4D.}

\noindent\mt{\textbf{Scoring.}}
We evaluate all models in a zero-shot manner without task-specific fine-tuning. Since the trained models learn a shared video-text embedding space, predictions for classification and retrieval tasks are performed by computing cosine similarity between the video (text) embedding 
and text (video) embeddings, selecting the most similar label for classification 
or ranking candidates for retrieval.
\mt{For video MLLMs that produce free-form text, we rank all verb labels by their length-normalized log-likelihood given the video and the target object, avoiding free-form generation and answer parsing.}

\begin{table*}[t]
\centering
\renewcommand{\arraystretch}{}

\resizebox{\linewidth}{!}{
\begin{tabular}{llccccccccc}
\toprule
\multirow{3}{*}{Model} 
& \multirow{3}{*}{Training Video Data}
& \multirow{3}{*}{Backbone}
& \multirow{3}{*}{\#Param}
& \multirow{3}{*}{TFLOPs}
& \multicolumn{2}{c}{Hand-Centric Video}
& \multicolumn{2}{c}{Object-Centric Video}
& \multicolumn{2}{c}{Original Video} \\
& & & & & \multicolumn{2}{c}{(Object Inpainted)}
& \multicolumn{2}{c}{(Hand Inpainted)}
& \multicolumn{2}{c}{} \\
\cmidrule(lr){6-7}
\cmidrule(lr){8-9}
\cmidrule(lr){10-11}
& & & & & Top1 & Top5 & Top1 & Top5 & Top1 & Top5 \\
\midrule
\multicolumn{11}{l}{\textit{Dual-encoder models}} \\
\addlinespace[3pt]
LaViLa~\cite{zhao2023learning}
& \egoicon EgoClip & TimeSformer & 528M & 3.5 & 12.7 {\small (-5.8)} & 30.3 {\small (-12.5)} 
& 10.3 {\small (-8.2)} & 22.9 {\small (-19.9)} 
& 18.5 & 42.8 \\
Helping Hands~\cite{zhang2023helping}
& \egoicon EgoClip w/ BBox & TimeSformer & 566M & 3.5 & 15.7 {\small (-8.1)} & 33.5 {\small (-14.9)}
& 14.2 {\small (-9.6)} & 28.3 {\small (-20.1)}
& 23.8 & 48.4 \\
\addlinespace[6pt]
{\color{black!60}
EgoVideo~\cite{pei2025modeling}}
& {\color{black!60}\makecell[l]{\egoicon EgoClip w/ more text,\\\exoicon KMash, WebVid10M, InternVid1\&2\\
}}
& {\color{black!60}InternVideo2}
& {\color{black!60}1.3B} & {\color{black!60}12.1}
& {\color{black!60}14.1 {\small (-8.8)}} & {\color{black!60}31.0 {\small (-15.1)}}
& {\color{black!60}16.4 {\small (-6.5)}} & {\color{black!60}35.1 {\small (-11.0)}}
& {\color{black!60}22.9} & {\color{black!60}46.1} \\
\addlinespace[6pt]
\textbf{Ours}
& \egoicon EgoClip w/ BBox & TimeSformer & 566M & 3.5 & \textbf{19.2} {\small (-7.7)} & \textbf{42.4} {\small (-12.9)}
& \textbf{19.5} {\small (-7.4)} & \textbf{38.8} {\small (-16.5)}
& \textbf{26.9} & \textbf{55.3} \\
\midrule
\multicolumn{11}{l}{\textit{LLM-integrated models (video MLLMs)}} \\
\addlinespace[3pt]
{\color{black!60}
VideoChat2~\cite{li2024mvbench}}
& {\color{black!60}\makecell[l]{VideoChat2-IT (1.9M, 34 sources,\\
incl.\ \egoicon\textbf{Ego4D 7.8K} as EgoQA), \exoicon +3 more}}
& {\color{black!60}\makecell[l]{InternVideo2\\ + Mistral-7B}}
& {\color{black!60}7.7B} & {\color{black!60}4.8}
& {\color{black!60}15.2 {\small (-6.4)}} & {\color{black!60}38.1 {\small (-10.5)}}
& {\color{black!60}19.2 {\small (-2.4)}} & {\color{black!60}42.4 {\small (-6.3)}}
& {\color{black!60}21.6} & {\color{black!60}48.6} \\
\addlinespace[6pt]
{\color{black!60}
LLaVA-Video-7B~\cite{zhang2024llava}}
& {\color{black!60}\makecell[l]{LLaVA-Video-178K\\
(incl.\ \egoicon\textbf{Ego4D 7.5K}), \exoicon +4 more}}
& {\color{black!60}\makecell[l]{SigLIP\\ + Qwen2-7B}}
& {\color{black!60}8.0B} & {\color{black!60}52.6}
& {\color{black!60}21.0 {\small (-5.9)}} & {\color{black!60}42.7 {\small (-9.6)}}
& {\color{black!60}24.3 {\small (-2.6)}} & {\color{black!60}48.5 {\small (-3.7)}}
& {\color{black!60}26.9} & {\color{black!60}52.3} \\
\bottomrule
\end{tabular}
}
\caption{Zero-shot evaluation on original and inpainted videos. Numbers in parentheses indicate absolute performance drops compared to the original video. \textbf{Bold} denotes the best score among the dual-encoder models. We report EgoVideo~\cite{pei2025modeling} and the LLM-integrated models in {\color{black!60}gray} since they are trained on much larger exocentric data. \egoicon\ and \exoicon\ denote egocentric and exocentric (third-person) video data, respectively. \mt{For MLLMs, we list the Ego4D-derived training data and denote the rest as \emph{+N more}. TFLOPs are measured for a single 16-frame forward pass at the DEHOI inference setting.} 
}
\label{tab:zero_shot_dehoi}
\vspace{-1em}
\end{table*}

\subsection{Results on DEHOI}
\label{sec:experiments::dehoi}
\subsubsection{Comparison with Baselines}
Table~\ref{tab:zero_shot_dehoi} compares egocentric dual-encoder VLMs and LLM-based video MLLMs with our model on DEHOI.
\mt{Most existing models suffer performance drops on hand- and object-centric videos compared with the original videos, consistent with the human study in \cref{sec:ci-hoi}. We further verify that these drops primarily result from cue removal rather than inpainting artifacts (see \cref{tab:artifact}). Importantly, the magnitude of the drop is asymmetric across models, revealing different reliance on hand and object cues.

LaViLa and Helping Hands, both trained on EgoClip, degrade more in the object-centric setting than in the hand-centric setting, indicating stronger reliance on hand cues. This asymmetry reverses for models trained on more diverse data: EgoVideo and the MLLMs suffer larger drops when objects are removed, suggesting greater reliance on object cues. 
In contrast, our model shows more comparable drops under hand and object removal in top-1 accuracy, with a reduced gap also in top-5 accuracy, unlike the other EgoClip-trained models.

In absolute performance, our model matches or outperforms all baselines on original videos and all but LLaVA-Video-7B in the hand-centric setting, despite being trained on a substantially smaller and less diverse corpus with up to $14\times$ fewer parameters and $15\times$ fewer FLOPs. In the object-centric setting, however, EgoVideo and the MLLMs perform comparably to or better than our model. This suggests that larger and more diverse training data, including non-egocentric videos and text supervision, can strengthen object-centric representations, although the gain over our model stays modest relative to this scale gap ($+4.8$ top-1 for LLaVA-Video-7B at $14\times$ the parameters and $15\times$ the FLOPs). However, this advantage does not extend to hand-centric or original videos: none surpasses our model on the original videos, and only LLaVA-Video-7B does so in the hand-centric setting. These results suggest that improving object understanding alone is insufficient; explicitly learning both hand and object dynamics, as in our approach, is key to robust performance across all video settings.}

\begin{figure}[t]
    \centering
    \includegraphics[width=\linewidth]{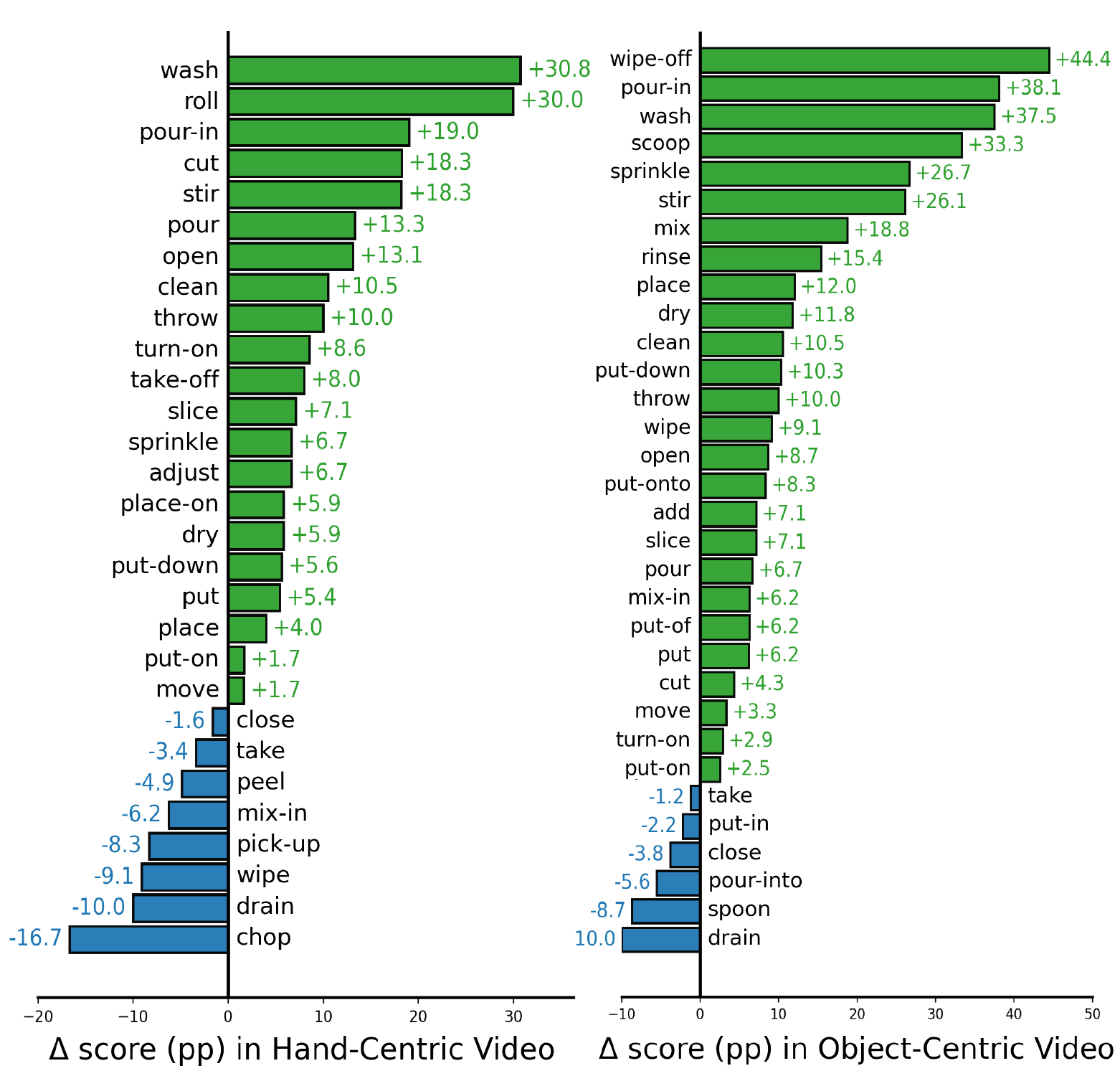}
    \vspace{-10pt}
    \caption{Verb-wise performance \textcolor{OliveGreen}{improvement} (\textcolor{Blue}{degradation}) of our model compared to Helping Hands~\cite{zhang2023helping} under hand-centric (left) and object-centric (right) settings. Verb classes with more than 10 samples and difference greater than 1\% are shown.}
    \label{fig:verb-wise_analysis}
    \vspace{-1em}
\end{figure}

\begin{figure}[t]
    \centering
    \includegraphics[width=\linewidth]{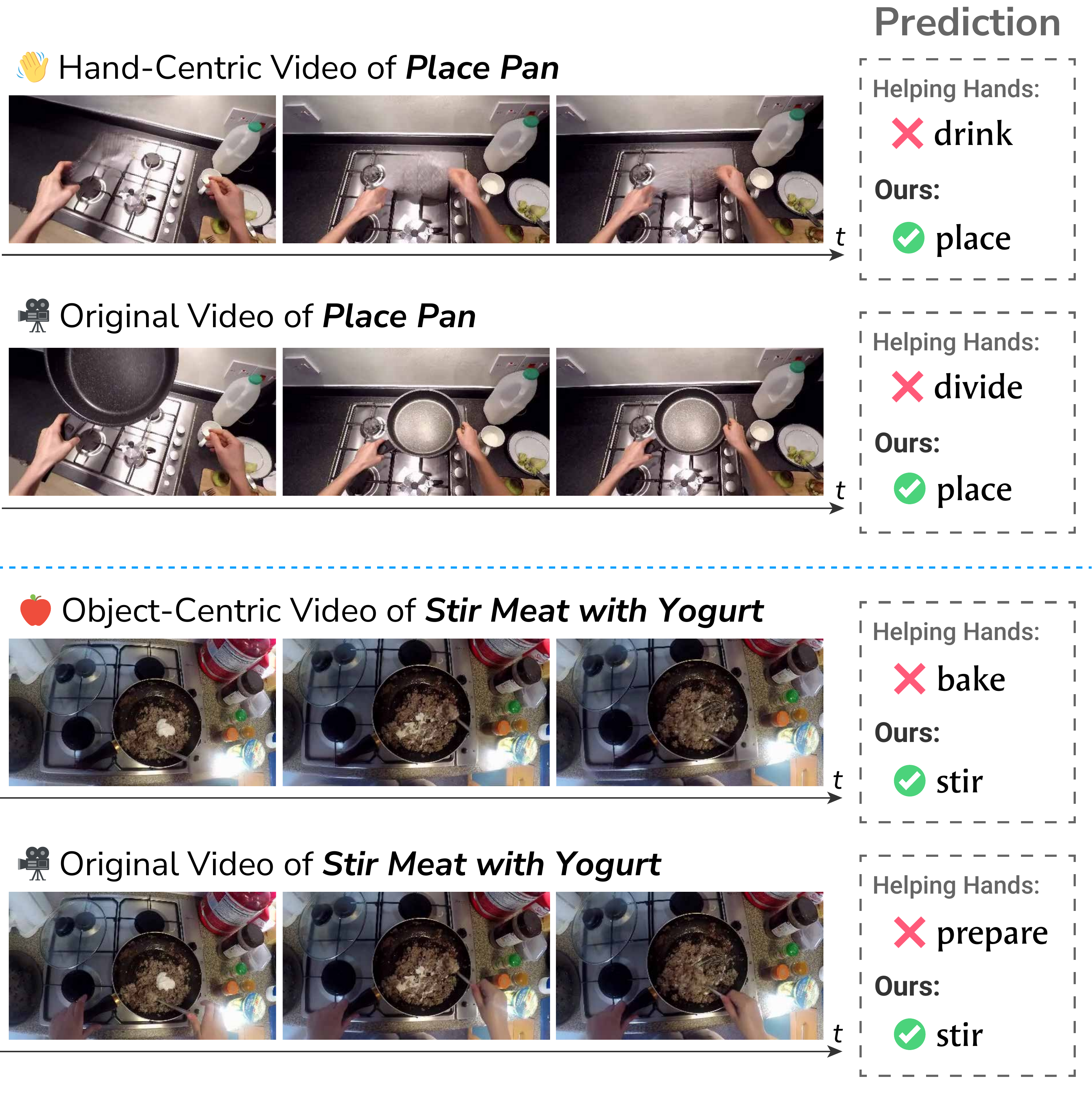}
    \vspace{-20pt}
    \caption{Qualitative results on cue-specific videos from DEHOI with results on original videos. Our model recognizes underlying dynamics from either hand or object observations, whereas existing models rely on spurious hand/object presence.}
    \label{fig:qualitative}
\end{figure}

\noindent
\textbf{Per-category analysis and qualitative results.}
\mt{
\cref{fig:verb-wise_analysis} compares our approach with Helping Hands for verbs with more than 10 samples and absolute differences above 1\%.
Our model improves many verbs, particularly those with distinctive hand motions or object transformations, such as \emph{wash}, \emph{pour-in}, and \emph{stir}.
For verbs with less distinctive hand motion, such as \emph{wipe-off}, \emph{mix}, and \emph{rinse}, gains are larger in the object-centric setting, suggesting improved use of object dynamics.
Performance decreases mainly for relatively static actions such as \emph{drain}, or contact-dominated actions such as \emph{take}.
The qualitative examples in~\cref{fig:qualitative} further illustrate this behavior.
Our model recognizes \emph{stir} from object appearance changes (bottom) and correctly predicts actions from hand-centric cues (top), while Helping Hands tends toward context-driven predictions such as \emph{bake}.
These results suggest that explicitly modeling interaction dynamics reduces reliance on static contextual correlations and encourages reasoning from object dynamics and hand motion.
}

\begin{table}[t]
\centering
\begin{subtable}[t]{\linewidth}
\centering
\resizebox{\linewidth}{!}{
\begin{tabular}{lcccccc}
\toprule
Method
& \multicolumn{2}{c}{Hand-Centric}
& \multicolumn{2}{c}{Object-Centric}
& \multicolumn{2}{c}{Original} \\
\cmidrule(lr){2-3}\cmidrule(lr){4-5}\cmidrule(lr){6-7}
& Top1 & Top5 & Top1 & Top5 & Top1 & Top5 \\
\midrule
w/o masked training & 18.0 & 40.7 & 16.9 & 33.6 & 26.0 & 54.2 \\
w/o dynamics semantics & 16.3 & 34.1 & 16.8 & 32.1 & 24.6 & 48.1 \\
Ours & \textbf{19.2} & \textbf{42.4} & \textbf{19.5} & \textbf{38.8} & \textbf{26.9} & \textbf{55.3} \\
\bottomrule
\end{tabular}
}
\vspace{-2pt}
\caption{Impact of core components.}
\label{tab:ablation::a}
\end{subtable}

\vspace{6pt}

\begin{subtable}[t]{\linewidth}
\centering
\resizebox{\linewidth}{!}{
\begin{tabular}{lcccccc}
\toprule
Masking Ratio
& \multicolumn{2}{c}{Hand-Centric}
& \multicolumn{2}{c}{Object-Centric}
& \multicolumn{2}{c}{Original} \\
\cmidrule(lr){2-3}\cmidrule(lr){4-5}\cmidrule(lr){6-7}
& Top1 & Top5 & Top1 & Top5 & Top1 & Top5 \\
\midrule
75\% & 19.0 & 41.6 & 18.6 & 36.2 & 26.0 & 53.3 \\
50\% & \textbf{19.2} & \textbf{42.4} & \textbf{19.5} & \textbf{38.8} & \textbf{26.9} & \textbf{55.3} \\
\bottomrule
\end{tabular}
}
\vspace{-2pt}
\caption{Ablation on masking ratio.}
\label{tab:ablation::b}
\end{subtable}

\vspace{6pt}

\begin{subtable}[t]{\linewidth}
\centering
\resizebox{\linewidth}{!}{
\begin{tabular}{lcccccc}
\toprule
Masking method
& \multicolumn{2}{c}{Hand-Centric}
& \multicolumn{2}{c}{Object-Centric}
& \multicolumn{2}{c}{Original} \\
\cmidrule(lr){2-3}\cmidrule(lr){4-5}\cmidrule(lr){6-7}
& Top1 & Top5 & Top1 & Top5 & Top1 & Top5 \\
\midrule
Random & 18.4 & 41.6 & 18.4 & 37.2 & 26.0 & 55.0 \\
Hand-object aware (Ours) & \textbf{19.2} & \textbf{42.4} & \textbf{19.5} & \textbf{38.8} & \textbf{26.9} & \textbf{55.3} \\
\bottomrule
\end{tabular}
}
\vspace{-2pt}
\caption{Ablation on masking strategy.}
\label{tab:ablation::e}
\end{subtable}

\vspace{-10pt}
\caption{Ablation of different modules across three video settings.}
\label{tab:ablation}
\vspace{-1em}
\end{table}

\subsubsection{Ablation}
\begin{figure}[t]
    \centering
    \includegraphics[width=1\linewidth]{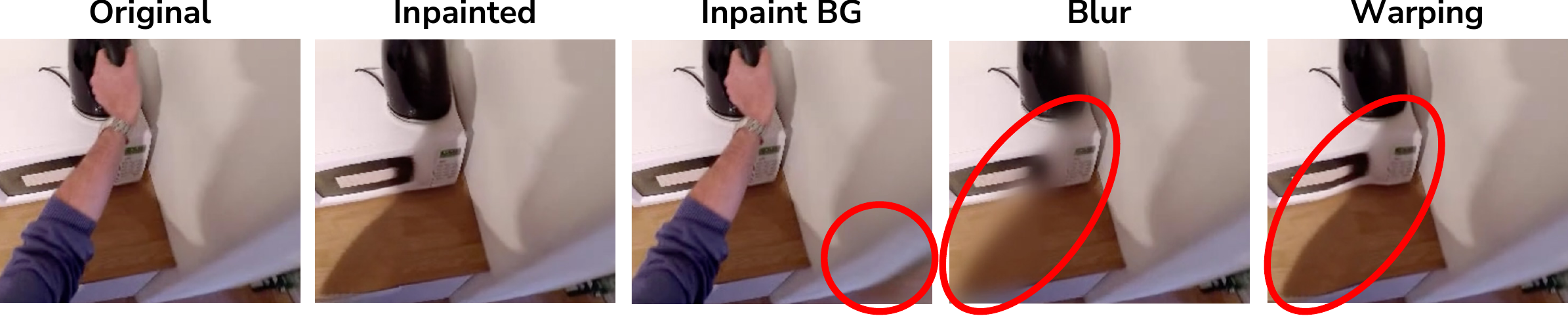}
    \vspace{-2em}
    \caption{Examples of the artifacts added for the analysis in \cref{tab:artifact}.}
    \label{fig:artifacts}
    \vspace{-5px}
\end{figure}

We ablate our components in~\cref{tab:ablation}.
In~\ref{tab:ablation::a}, HDA's dynamics prediction substantially improves both hand- and object-centric recognition, while hand-object masked training provides stronger gains in object-centric recognition.
Together, they consistently improve performance across cue-specific and original videos.
In~\ref{tab:ablation::b}, we investigate the effect of different masking ratios. A 50\% masking ratio outperforms 75\%, suggesting that moderate masking removes semantically important regions while preserving sufficient context to model relationships among visible patches.
We further compare our masking strategy with random masking in~\ref{tab:ablation::e}. Hand-object masking consistently performs better across video types, demonstrating the benefit of entity-aware masking for robust interaction representations. Ablations on masking probabilities and additional trainable parameters are in \cref{subsec:ablation_supp}.
\mt{To analyze the impact of HOI detection, we perturbed the hand-object bounding boxes by (1) shifting them with a mean magnitude of 14 / 28 px (1--2 patches) and (2) randomly dropping detections ($p=0.5$).
Results in \cref{tab:hoi-detection} show that the method reflects detection quality, indicating that hand-object localization is important for modeling HOI cues.}
Since inpainting may introduce artifacts into videos, we also evaluate their effect on CI-HOI. We test multiple settings, including background inpainting and additional blur/warping artifacts in the inpainted hand/object regions (\cref{fig:artifacts}). As shown in \cref{tab:artifact}, performance remains similar across settings, suggesting that the degradation mainly comes from missing hand/object cues, not inpainting artifacts.

\begin{table}[t]
\centering
\resizebox{\linewidth}{!}{
\setlength{\tabcolsep}{10pt}
\begin{tabular}{lcccccc}
\toprule
Method
& \multicolumn{2}{c}{Hand-Centric}
& \multicolumn{2}{c}{Object-Centric}
& \multicolumn{2}{c}{Original} \\
\cmidrule(lr){2-3}\cmidrule(lr){4-5}\cmidrule(lr){6-7}
& Top1 & Top5 & Top1 & Top5 & Top1 & Top5 \\
\midrule
Ours & \textbf{19.2} & \textbf{42.4} & \textbf{19.5} & \textbf{38.8} & \textbf{26.9} & \textbf{55.3} \\
+ Shift bbox ($14$px) & 19.0 & 41.8 & 19.1 & 37.8 & 26.4 & 55.2 \\
+ Shift bbox ($28$px) & 19.0 & 40.8 & 18.8 & 37.3 & 25.9 & 54.0 \\
+ Drop bbox ($p=0.5$) & 19.2 & 41.8 & 18.9 & 38.1 & 26.3 & 55.0 \\
\bottomrule
\end{tabular}
}
\vspace{-4pt}
\caption{\mt{Impact of HOI-detection quality. Shifts are reported as the mean displacement magnitude of the boxes.}}
\label{tab:hoi-detection}
\end{table}

\begin{table}[t]
    \centering
    \resizebox{\linewidth}{!}{
    \setlength{\tabcolsep}{10pt}
    \begin{tabular}{lcccccc}
    \toprule
    Method
    & \multicolumn{2}{c}{Hand-Centric}
    & \multicolumn{2}{c}{Object-Centric}
    & \multicolumn{2}{c}{Original} \\
    \cmidrule(lr){2-3}\cmidrule(lr){4-5}\cmidrule(lr){6-7}
    & Top1 & Top5 & Top1 & Top5 & Top1 & Top5 \\
    \midrule
    DEHOI (our testbed) & \textbf{19.2} & \textbf{42.4} & \textbf{19.5} & \textbf{38.8} & \textbf{26.9} & \textbf{55.3} \\
    Inpainting background & -- & -- & -- & -- & 26.1 & 54.9 \\
    Blur inpainted area  & 18.7 & 41.2 & 19.6 & 38.9 & -- & -- \\
    Warp inpainted area  & 19.4 & 42.4 & 19.5 & 38.8 & -- & -- \\
    \bottomrule
    \end{tabular}
    }
\caption{Effect of artifacts on our model's performance in CI-HOI.}
\label{tab:artifact}
\end{table}

\subsection{Results on Object State Recognition}
\label{sec:experiments::osr}

\noindent
STATUS Bench evaluates recognition of object states and their changes.
We compare our model with LaViLa and Helping Hands, which are trained only on egocentric videos as in our setting, to isolate the effect of our training framework.
As shown in~\cref{tab:status}, our model outperforms or matches these egocentric-only baselines on all tasks. A notable gain is observed on State Change Identification, which requires identifying dynamic transitions between object states rather than static states in isolation. This improvement further affirms the effectiveness of our explicit modeling of object dynamics.

\begin{table}[t]
\centering
\resizebox{\linewidth}{!}{
\begin{tabular}{lccc}
\toprule
Method
& \makecell{Object State\\Identification}
& \makecell{Image\\Retrieval}
& \makecell{State Change\\Identification} \\
\midrule
LaViLa~\cite{zhao2023learning} & 49.1 & 50.5 & 24.7 \\
Helping Hands~\cite{zhang2023helping} & 48.4 & \textbf{50.7} & 26.5 \\
\midrule
\textbf{Ours} & \textbf{49.4} & \textbf{50.7} & \textbf{30.0} \\
\bottomrule
\end{tabular}
}
\vspace{-4pt}
\caption{\mt{Results on STATUS Bench~\cite{ukai2025status}.}}
\label{tab:status}
\vspace{-1em}
\end{table}

\section{Conclusion}
We proposed a new training paradigm that exploits complementary hand- and object-centric cues through hand-object masked training and a decoder that facilitates semantic alignment of cue-specific dynamics.
Our Cue-Isolated HOI evaluation shows that, unlike existing egocentric VLMs that rely more heavily on hand-centric cues, our training reduces this over-reliance and improves action recognition accuracy.
We hope these findings provide a useful direction toward more holistic semantic understanding of HOI.

\section*{LLM Usage Statement}
We used LLMs (Claude and ChatGPT) for coding assistance and language editing. The research ideas, experimental design, and conclusions are our own. We reviewed all LLM-assisted content, verified the code and reported numbers against our experimental results, and take full responsibility for the manuscript.

{
    \small
    \bibliographystyle{ieeenat_fullname}
    \bibliography{main}
}

\clearpage
\setcounter{page}{1}
\maketitlesupplementary

\appendix
\crefalias{section}{appendix}
\crefalias{subsection}{appendix}
\crefalias{subsubsection}{appendix}

\setcounter{section}{0}
\setcounter{subsection}{0}

\renewcommand{\thesection}{\Alph{section}}
\renewcommand{\thesubsection}{\thesection.\arabic{subsection}}

\section{Training Data Preparation}
\noindent
\textbf{Hand-object bounding box.}
Our proposed training strategy assumes localized bounding boxes for hands and manipulated objects, which are used both for hand-object masked training (\cref{subsec:hand-object_masked_training}) and for training the HDA decoder (\cref{subsec:hda_decoder}). In our experiments, we used precomputed detection results from the 100DOH hand-object detector~\cite{shan2020understanding} provided by~\cite{zhang2023helping}. Following~\cite{zhang2023helping}, we use the top two hands and top four manipulated objects detected in EgoClip~\cite{lin2022egocentric} videos.

\noindent
\textbf{Noun and verb.}
Extracted noun and verb annotations for each narration are provided in EgoClip~\cite{lin2022egocentric}. Each narration may contain multiple nouns and verbs. The loss computation for cases involving multiple nouns or verbs is described in the following objectives section.

\section{Objectives}
\subsection{Bounding Box Loss}
As described in \cref{subsec:hda_decoder}, we apply $\ell_1$ loss and GIoU loss to the Hungarian-matched predictions. However, some unmatched predictions may arise from the object-centric embeddings. Following \cite{zhang2023helping}, we do not penalize these unmatched predictions, unlike in traditional detection tasks, since the prepared object bounding boxes may be missing.

\subsection{Noun- and Verb-Level NCE Loss}
We use the text embeddings of the extracted nouns and verbs from the narration as positive supervision signals. 
For each semantic embedding $\mathbf{s}$ (\cref{subsec:hda_decoder}), we compute an InfoNCE loss for nouns and verbs against the corresponding taxonomy dictionary provided by Ego4D.
For noun supervision, each object-centric embedding is aligned with its corresponding noun in the narration. 
Since each object-centric embedding corresponds to a single noun label, we use a standard single-positive InfoNCE objective:
\begin{equation}
\mathcal{L}_{\text{Noun}}
=
-\frac{1}{K}
\sum_{j=1}^{K}
\log
\frac{
\exp({\mathbf{s}^o_j}^\top \mathbf{t}_{\text{noun}}^{(\sigma(j))} / \tau)
}{
\sum_{k \in \mathcal{D}_{\text{noun}}}
\exp({\mathbf{s}^o_j}^\top \mathbf{t}^{\text{noun}}_k / \tau)
}.
\end{equation}
In contrast, verb supervision is inherently multi-label. 
A narration may contain multiple verbs, and each semantic embedding should be compatible with all verbs appearing in the narration. 
Therefore, we adopt a multi-positive InfoNCE formulation where all ground-truth verbs contribute as positives:

\begin{equation}
\begin{aligned}
\mathcal{L}_{\text{Verb}} =
&-\frac{1}{N_h}\sum_{i=1}^{N_h}
\log
\frac{
\sum_{p\in\mathcal{P}}
\exp(\mathbf{s}_i^{h\top}\mathbf{t}^{\text{verb}}_{p}/\tau)
}{
\sum_{k\in\mathcal{D}_{\text{verb}}}
\exp(\mathbf{s}_i^{h\top}\mathbf{t}^{\text{verb}}_{k}/\tau)
} \\
&-\frac{1}{K}\sum_{j=1}^{K}
\log
\frac{
\sum_{p\in\mathcal{P}}
\exp(\mathbf{s}_j^{o\top}\mathbf{t}^{\text{verb}}_{p}/\tau)
}{
\sum_{k\in\mathcal{D}_{\text{verb}}}
\exp(\mathbf{s}_j^{o\top}\mathbf{t}^{\text{verb}}_{k}/\tau)
}.
\end{aligned}
\end{equation}

Here, $\mathbf{s}_i^h$ and $\mathbf{s}_j^o$ denote the $i$-th hand-centric and $j$-th object-centric embedding, respectively. 
$N_h$ and $K$ represent the numbers of hand- and object-centric embeddings. 
$\mathbf{t}^{\text{noun}}_k$ and $\mathbf{t}^{\text{verb}}_k$ denote noun and verb embeddings from the Ego4D taxonomy dictionaries $\mathcal{D}_{\text{noun}}$ and $\mathcal{D}_{\text{verb}}$. 
$\sigma(j)$ maps the $j$-th object-centric embedding to its corresponding noun label. 
$\mathcal{P}$ denotes the set of verb indices appearing in the narration, and $\tau$ is a temperature parameter.

\section{DEHOI}
\subsection{Construction}
\subsubsection{Data Source.}
We source HOI videos from EPIC-KITCHENS-100 (EK-100)~\cite{damen2022rescaling}, which contains around 100 hours of unscripted egocentric cooking videos recorded in 45 kitchen environments. EK-100 provides temporally localized action segments with annotated start/end times and narrations, where narrations are mapped to verb-noun pairs (97 verb classes and 300 noun classes). We use the validation split for our benchmark since the test split is not publicly available.
We use the pixel-level segmentation masks and hand-object contact relations from EPIC-KITCHENS VISOR~\cite{darkhalil2022epic}, which provides hand and active-object regions required for inpainting. VISOR includes both manually-annotated temporally sparse masks and dense masks obtained by interpolating the sparse annotations for additional frames.

\subsubsection{Prepare Masks.}
Since VISOR does not provide segmentation masks aligned with every action annotation in EK-100, we retrieve the available hand and object masks in three steps:
\begin{itemize}
    \item \textbf{Active object retrieval.}
    We treat the annotated noun of each action in EK-100 as the active object.
    For each frame within the action segment, we retrieve segmentation masks corresponding to the annotated noun as well as the \emph{left hand} and \emph{right hand} from VISOR.
    
    \item \textbf{Fallback via contact annotations.}
    If no segmentation mask is available in VISOR for the active object, we inspect the temporally sparse hand-object contact annotations in VISOR within the action segment.
    When an object is annotated as being in contact with either hand, we retrieve the corresponding sparse object mask.
    
    \item \textbf{Filtering.}
    If neither dense nor sparse annotations are available for the active object, we discard the action segment from further analysis.
\end{itemize}

After retrieving the available masks, we propagate them to unannotated video frames using SAM2~\cite{ravi2024sam} for the following inpainting process.

\subsection{Statistics}
\begin{figure}[t]
    \centering
    \includegraphics[width=1\linewidth]{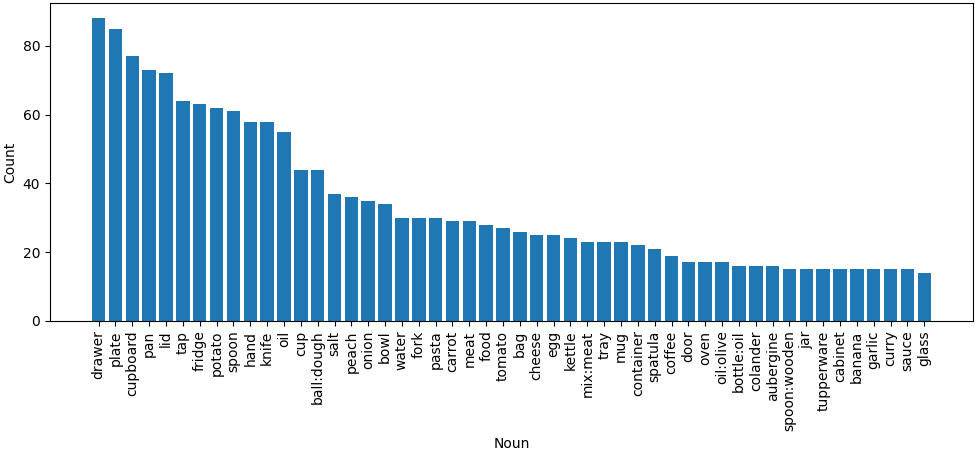}
    \caption{Top-50 noun distribution in DEHOI.}
    \label{fig:noun_distribution}
\end{figure}

\begin{figure}[t]
    \centering
    \includegraphics[width=1\linewidth]{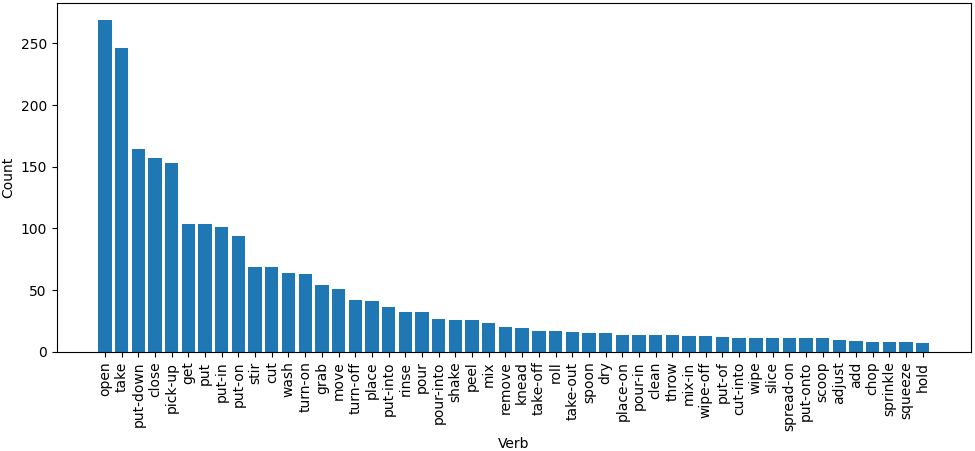}
    \caption{Top-50 verb distribution in DEHOI.}
    \label{fig:verb_distribution}
\end{figure}

DEHOI consists of 2,652 videos for each of the hand- and object-centric settings.
\cref{fig:noun_distribution} and \cref{fig:verb_distribution} show the distributions of the top-50 most frequent annotated nouns and verbs in the DEHOI videos.

\section{Human Study}
\label{sec:human_study_supp}
We provide the interface and the procedure of the human study reported in \cref{sec:ci-hoi}.

\subsection{Interface}
\begin{figure}[t]
    \centering
    \includegraphics[width=0.95\linewidth]{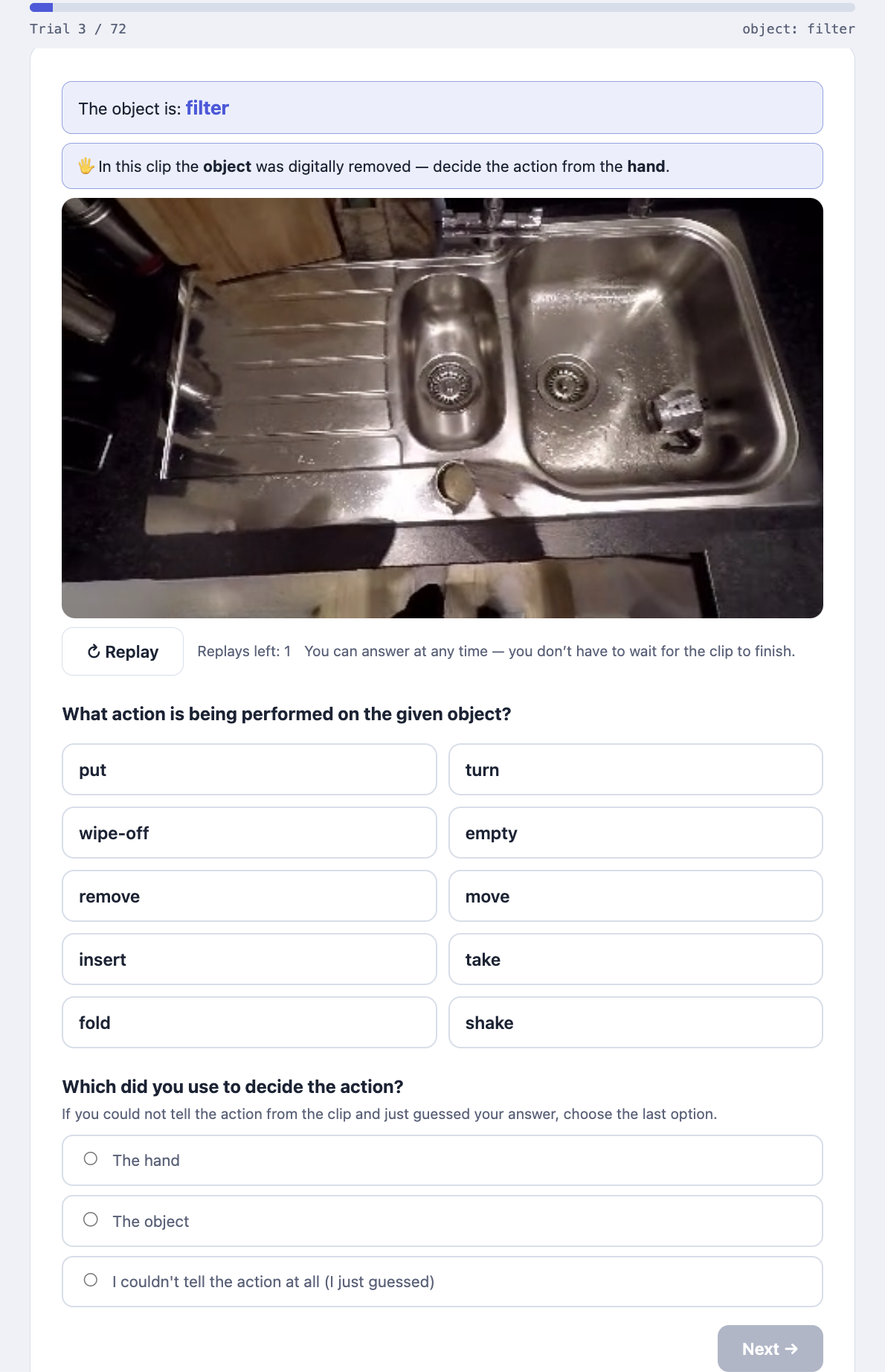}
    \caption{Interface of the human study, showing a trial of the hand-centric condition. The participant is given the object name and a banner describing which cue was removed, watches the clip with at most one replay, selects the action from ten candidate verbs, and reports which cue was used to decide the action.}
    \label{fig:human_study_ui}
\end{figure}

\cref{fig:human_study_ui} shows the web interface used to collect the responses. Each screen presents a single trial and consists of the following elements.

\noindent\textbf{Trial header.}
A progress bar and a trial counter indicate how far the participant has advanced through the session.

\noindent\textbf{Target object.}
The name of the manipulated object (\eg, \emph{filter}) is shown above the clip. This follows the protocol used for the models, which also receive the object name, and keeps the task focused on recognizing the interaction rather than identifying the object.

\noindent\textbf{Condition banner.}
A banner states which cue has been removed from the clip, \eg, ``In this clip the object was digitally removed --- decide the action from the hand'' for the hand-centric condition. Participants therefore know which cue has been removed.

\noindent\textbf{Clip playback.}
The clip is played without audio. Participants may replay it once and can answer at any time without waiting for the playback to finish. Limiting the number of replays keeps the viewing time comparable across participants while still allowing a second look at these short clips.

\noindent\textbf{Verb question.}
Participants choose the action from ten candidate verbs, consisting of the ground-truth verb of the clip and nine distractors, giving a chance level of 10\%. The distractors are the nine verb classes that most frequently co-occur with the target object in the EK-100 training set, excluding classes that share a word with the ground-truth verb (\eg, \emph{put} for \emph{put-down}). The candidates are presented in a fixed two-column layout, and the trial can be submitted only after this question is answered.

\noindent\textbf{Cue-source question.}
In the two cue-isolated conditions, participants additionally report which cue they used to decide the action: \emph{the hand}, \emph{the object}, or \emph{I couldn't tell the action at all (I just guessed)}. The last option separates trials in which the verb was genuinely recovered from the visible cue from trials in which the participant guessed, which we report in \cref{tab:human_study}.

\subsection{Procedure}
Each participant completes all 96 trials, 32 under each of the three conditions. A clip is presented to a given participant under only one condition, so that a participant never sees the same interaction twice. Conditions are assigned with a Latin square, so that each clip receives exactly six responses under each condition. Across the 18 participants this yields 576 responses per condition. Only completed sessions are analyzed. Participants receive no feedback about correctness during the study.

\subsection{Verb-wise Results}
\label{subsec:human_verbwise}
\begin{figure}[t]
\centering
\includegraphics[width=\linewidth]{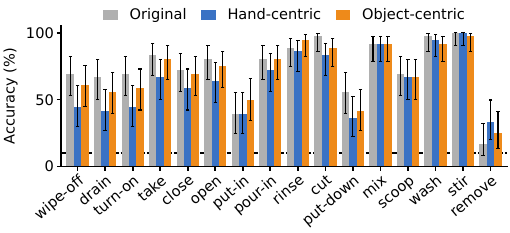}
\vspace{-10pt}
\caption{\mt{Verb-wise human accuracy under the three conditions, sorted by the gap between the object- and hand-centric conditions. Error bars denote 95\% Wilson confidence intervals ($n{=}36$ per bar), and the dashed line denotes the chance level (10\%).}}
\label{fig:human_verbwise}
\end{figure}

\cref{fig:human_verbwise} shows the verb-wise human accuracy.
Object-centric accuracy is higher or comparable for all verbs except \emph{remove}.
Simple hand motions provide insufficient cues for verbs such as \emph{take}, \emph{put-down}, \emph{pour-in}, and \emph{drain}, whereas distinctive hand motions make verbs such as \emph{stir}, \emph{wash}, and \emph{mix} recognizable from hand cues alone.

\section{Details on Experiments}
\subsection{LoRA Configuration}
\label{subsec:lora_supp}
In the video encoder backbone, LoRA is applied to the spatial and temporal self-attention modules, specifically to the query, key, value, and output projection layers.
In the HDA decoder, LoRA is inserted into the multi-head attention and feed-forward blocks, including their projection layers.
The LoRA rank is set to 8 and the scaling factor $\alpha$ to 4.

\subsection{Evaluation Protocol}
When evaluating on \textbf{DEHOI} and \textbf{DROID}, the task is verb classification given the target object category and the video. Since all baselines are trained using video-language contrastive learning, which produces a joint embedding space for videos and language, predictions are made by computing the cosine similarity between the video embedding and the text embeddings of all possible interaction descriptions for the target object, and selecting the most similar label. We construct negative textual descriptions by replacing the verb in the positive description with other verb classes taken from the EPIC-KITCHENS verb taxonomy~\cite{damen2022rescaling}.
When evaluating on \textbf{STATUS Bench}, we use state description or state-change description provided in the benchmark to conduct image/text retrieval and multiple-choice question task.
\mt{For \textbf{DROID}, we use 16 input frames, following the protocol used for DEHOI.}

\subsection{Ablation}
\label{subsec:ablation_supp}
\cref{tab:ablation::lambda_nce} and \cref{tab:ablation::lambda_verb} present the ablation study on the loss weights of the EgoNCE loss ($\lambda_{\text{vt}}$) and the verb loss ($\lambda_{\text{verb}}$), respectively.
By default, we set the loss weights to $\lambda_{\text{vt}}=0.2$, $\lambda_{\text{verb}}=0.3$, and $\lambda_{\text{noun}}=0.5$.
In this ablation, we fix the noun loss weight and vary the other two loss weights.

As shown in \cref{tab:ablation::lambda_nce}, $\lambda_{\text{vt}}=0.2$ provides the best overall performance across the three evaluation settings.
For the verb loss, increasing the weight to $0.5$ slightly improves performance in the hand- and object-centric settings. However, since our primary goal is to improve performance in the original setting, we select $\lambda_{\text{verb}}=0.3$ in the main texts, which achieves the best Top-1 accuracy on the original videos.

\mt{\cref{tab:ablation::c} analyzes the impact of varying the hand/object masking probabilities.
As our masked training explicitly depends on hand and object locations, we examine their relative importance by adjusting the masking probabilities.
Increasing the probability of masking hand regions consistently degrades performance, indicating that hand-related cues play a crucial role in modeling interaction dynamics.
\cref{tab:ablation::d} analyzes the effect of varying the probability of background masking relative to hand/object masking.
We find that leaving the hand and object regions unmasked with a 20\% probability helps the model retain reasoning over hand-object relationships while still learning from partial observations for the remaining training samples.}

\mt{\noindent\textbf{Effect of additional trainable parameters.}
To verify that our gains do not simply come from these additional trainable parameters, we train a baseline (HH + LoRA) that starts from the same Helping Hands checkpoint and uses the same LoRA modules and the same number of training iterations as ours, but keeps the original Helping Hands objectives without our dynamics prediction and hand-object masked training.
As shown in \cref{tab:ablation::lora_only}, HH + LoRA does not improve over Helping Hands and even slightly degrades in all settings, whereas our model improves top-1 accuracy over HH + LoRA by $+4.9$, $+5.5$, and $+4.2$ points on hand-centric, object-centric, and original videos, respectively.
This confirms that our improvements stem from our training framework rather than from the increased number of trainable parameters.}

\begin{table}[t]
\centering
\begin{subtable}[t]{\linewidth}
\centering
\small
\setlength{\tabcolsep}{3pt}
\begin{tabular}{lcccccc}
\toprule
$\lambda_{\text{vt}}$
& \multicolumn{2}{c}{Hand-Centric}
& \multicolumn{2}{c}{Object-Centric}
& \multicolumn{2}{c}{Original} \\
\cmidrule(lr){2-3}\cmidrule(lr){4-5}\cmidrule(lr){6-7}
& Top1 & Top5 & Top1 & Top5 & Top1 & Top5 \\
\midrule
0.0 & 19.1 & \textbf{43.2} & 18.6 & 38.6 & 25.6 & 54.8 \\
0.2 & \textbf{19.2} & 42.4 & \textbf{19.5} & \textbf{38.8} & \textbf{26.9} & \textbf{55.3} \\
0.4 & 18.9 & 40.8 & 18.9 & 37.7 & 26.5 & 54.8 \\
\bottomrule
\end{tabular}
\caption{Ablation on $\lambda_{\text{vt}}$.}
\label{tab:ablation::lambda_nce}
\end{subtable}

\vspace{6pt}

\begin{subtable}[t]{\linewidth}
\centering
\small
\setlength{\tabcolsep}{3pt}
\begin{tabular}{lcccccc}
\toprule
$\lambda_{\text{Verb}}$
& \multicolumn{2}{c}{Hand-Centric}
& \multicolumn{2}{c}{Object-Centric}
& \multicolumn{2}{c}{Original} \\
\cmidrule(lr){2-3}\cmidrule(lr){4-5}\cmidrule(lr){6-7}
& Top1 & Top5 & Top1 & Top5 & Top1 & Top5 \\
\midrule
0.1 & 18.3 & 40.1 & 18.2 & 35.2 & 26.0 & 54.2 \\
0.3  & 19.2 & 42.4 & 19.5 & 38.8 & \textbf{26.9} & 55.3 \\
0.5 & \textbf{19.3} & \textbf{43.1} & \textbf{19.5} & \textbf{39.5} & 26.7 & \textbf{55.7} \\
\bottomrule
\end{tabular}
\caption{Ablation on $\lambda_{\text{Verb}}$.}
\label{tab:ablation::lambda_verb}
\end{subtable}
\caption{Ablation study on loss weights.}
\end{table}

\begin{table}[t]
\centering
\begin{subtable}[t]{\linewidth}
\centering
\small
\setlength{\tabcolsep}{3pt}
\begin{tabular}{lcccccc}
\toprule
Hand:Object
& \multicolumn{2}{c}{Hand-Centric}
& \multicolumn{2}{c}{Object-Centric}
& \multicolumn{2}{c}{Original} \\
\cmidrule(lr){2-3}\cmidrule(lr){4-5}\cmidrule(lr){6-7}
& Top1 & Top5 & Top1 & Top5 & Top1 & Top5 \\
\midrule
0.7:0.3 & 18.1 & 40.9 & 18.4 & 37.8 & 25.7 & 54.8 \\
0.5:0.5 & 18.1 & 41.5 & 19.0 & 38.9 & 25.8 & 54.7 \\
0.3:0.7 & 18.7 & 42.3 & 19.4 & \textbf{39.6} & 26.3 & 55.1 \\
0.1:0.9 & \textbf{19.2} & \textbf{42.4} & \textbf{19.5} & 38.8 & \textbf{26.9} & \textbf{55.3} \\
\bottomrule
\end{tabular}
\caption{Ablation on the hand/object masking probability.}
\label{tab:ablation::c}
\end{subtable}

\vspace{1em}

\begin{subtable}[t]{\linewidth}
\centering
\small
\setlength{\tabcolsep}{3pt}
\begin{tabular}{lcccccc}
\toprule
Background
& \multicolumn{2}{c}{Hand-Centric}
& \multicolumn{2}{c}{Object-Centric}
& \multicolumn{2}{c}{Original} \\
\cmidrule(lr){2-3}\cmidrule(lr){4-5}\cmidrule(lr){6-7}
Masking Balance & Top1 & Top5 & Top1 & Top5 & Top1 & Top5 \\
\midrule
0.6 & 18.8 & 41.6 & 18.7 & 38.0 & 26.4 & 55.3 \\
0.4 & 18.6 & 41.6 & 18.9 & 37.9 & 26.2 & \textbf{55.6} \\
0.2 & \textbf{19.2} & \textbf{42.4} & \textbf{19.5} & \textbf{38.8} & \textbf{26.9} & 55.3 \\
0.0 (entity-only) & 18.7 &  40.7 & 18.8 & 38.1 & 26.1 & 54.4 \\
\bottomrule
\end{tabular}
\vspace{-2pt}
\caption{Ablation on the background masking probability.}
\label{tab:ablation::d}
\end{subtable}
\caption{\mt{Ablation on the masking probabilities.}}
\label{tab:ablation::mask_balance}
\end{table}

\begin{table}[t]
\centering
\small
\setlength{\tabcolsep}{3pt}
\begin{tabular}{lcccccc}
\toprule
Method
& \multicolumn{2}{c}{Hand-Centric}
& \multicolumn{2}{c}{Object-Centric}
& \multicolumn{2}{c}{Original} \\
\cmidrule(lr){2-3}\cmidrule(lr){4-5}\cmidrule(lr){6-7}
& Top1 & Top5 & Top1 & Top5 & Top1 & Top5 \\
\midrule
Helping Hands~\cite{zhang2023helping} & 15.7 & 33.5 & 14.2 & 28.3 & 23.8 & 48.4 \\
HH + LoRA & 14.3 & 31.4 & 14.0 & 27.2 & 22.7 & 45.6 \\
\textbf{Ours} & \textbf{19.2} & \textbf{42.4} & \textbf{19.5} & \textbf{38.8} & \textbf{26.9} & \textbf{55.3} \\
\bottomrule
\end{tabular}
\vspace{-4pt}
\caption{\mt{Effect of additional trainable parameters on DEHOI. HH + LoRA is trained from Helping Hands with the same LoRA modules and training iterations as ours, but with the original Helping Hands objectives.}}
\label{tab:ablation::lora_only}
\end{table}

\subsection{More Qualitative Results}
\begin{figure}
    \centering
    \includegraphics[width=1\linewidth]{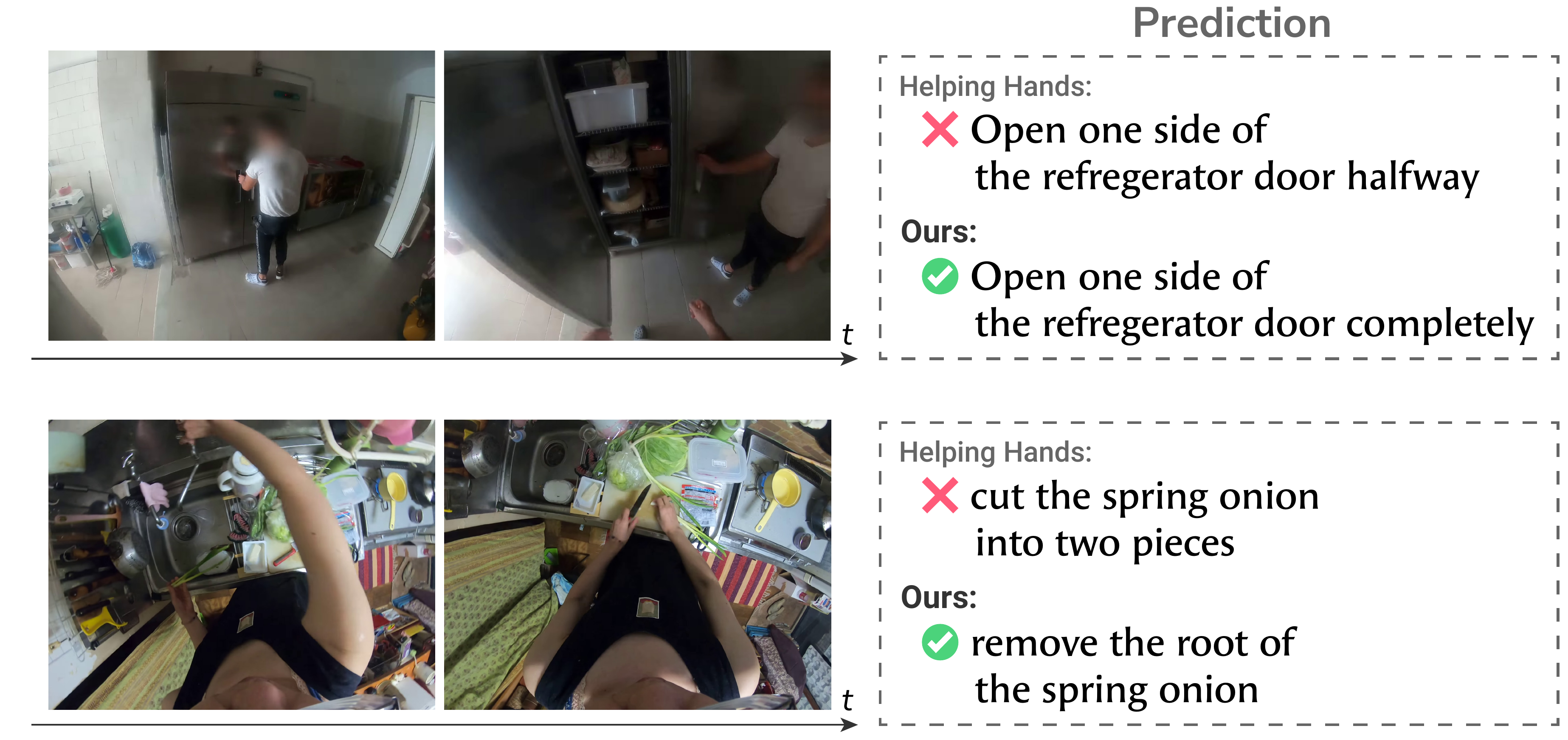}
    \caption{Qualitative results of State Change Identification task in STATUS Bench~\cite{ukai2025status}.}
    \label{fig:qualitative_status}
\end{figure}

\noindent\textbf{STATUS Bench.} As shown in \cref{sec:experiments::osr}, the model trained with our approach demonstrates clear improvements on the State Change Identification (SCI) task in STATUS Bench~\cite{ukai2025status}. We present qualitative results of the SCI task in \cref{fig:qualitative_status}. The SCI task requires recognizing subtle differences in object state changes described in text and selecting the description that best matches a given pair of images depicting a state change. The results show that our model captures object state changes in image pairs and correctly selects the appropriate description in both cases where the target object is large (top) and small (bottom).

\subsection{Comparison with Human Performance}
\label{subsec:human_correlation}
\mt{\noindent\textbf{Protocol.}
We evaluate models on the 96 clips of the human study (\cref{sec:ci-hoi}) under the same task setting.
Each clip is presented under the three conditions, and the model selects the verb out of the same 10 candidates that were shown to the participants (chance level 10\%).
Since participants are given the target object name, each candidate text is formed as ``\textless verb\textgreater\ \textless noun\textgreater'' (\eg, \emph{wipe off hand}), and the audio is removed.
Human accuracy is averaged over the 18 participants, whereas model accuracy is computed over the 96 clips.}

\mt{\noindent\textbf{Accuracy per condition.}
\cref{tab:human_vs_models} reports the accuracy under this setting.
Humans lose more accuracy when the object is removed than when the hands are removed (hand-centric $-9.7$ vs.\ object-centric $-3.1$).
In contrast, all models degrade most in the object-centric condition in this setting.
This differs from the DEHOI protocol, where our model reduces the gap between the hand- and object-centric conditions (see also \cref{tab:human_vs_models_97}).
We attribute this to the restricted candidate set: narrowing the full verb vocabulary down to ten candidates biases the task toward distinctions that hand cues can resolve more easily.
Even so, our model achieves the best accuracy among the models in every condition and the smallest drops from the original videos ($-3.1$ and $-11.4$).
Note that one clip corresponds to about one point of accuracy in this setting, so small differences should be interpreted with caution.}

\mt{\noindent\textbf{Effect of the candidate set.}
\cref{tab:human_vs_models_97} evaluates the same 96 clips using all 97 EPIC-KITCHENS-100 verb classes as candidates (chance level $\simeq1.0\%$), \ie, the protocol used for DEHOI.
This harder setting widens the gap between our model and the baselines: our model outperforms Helping Hands and LaViLa in every condition.
Our model also shows less accuracy gap between the hand- and object-centric conditions, whereas the baselines perform much worse in the object-centric condition than in the hand-centric condition.}

\begin{table}[t]
\centering
\small
\setlength{\tabcolsep}{4pt}
\renewcommand{\arraystretch}{1.05}
\begin{tabular}{lccc}
\toprule
Model & Original & Hand-Centric & Object-Centric \\
\midrule
Human (N=18) & 73.6 & 63.9 {\small (-9.7)} & 70.5 {\small (-3.1)} \\
\midrule
LaViLa~\cite{zhao2023learning} & 52.1 & 43.8 {\small (-8.3)} & 34.4 {\small (-17.7)} \\
Helping Hands~\cite{zhang2023helping} & 55.2 & 49.0 {\small (-6.2)} & 38.5 {\small (-16.7)} \\
\textbf{Ours} & \textbf{56.2} & \textbf{53.1} {\small (-3.1)} & \textbf{44.8} {\small (-11.4)} \\
\bottomrule
\end{tabular}
\vspace{-10pt}
\caption{\mt{Accuracy on the 96 clips of the human study under the same 10-way setting. Numbers in parentheses are the drops from the original videos. \textbf{Bold} denotes the best score among the models.}}
\label{tab:human_vs_models}
\end{table}

\begin{table}[t]
\centering
\small
\setlength{\tabcolsep}{2pt}
\renewcommand{\arraystretch}{1.05}
\begin{tabular}{lcccccc}
\toprule
\multirow{2}{*}{Model}
& \multicolumn{2}{c}{Original}
& \multicolumn{2}{c}{Hand-Centric}
& \multicolumn{2}{c}{Object-Centric} \\
\cmidrule(lr){2-3} \cmidrule(lr){4-5} \cmidrule(lr){6-7}
& Top1 & Top5 & Top1 & Top5 & Top1 & Top5 \\
\midrule
LaViLa~\cite{zhao2023learning} & 27.1 & 49.0 & 21.9 & 40.6 & 11.5 & 39.6 \\
Helping Hands~\cite{zhang2023helping} & 25.0 & 55.2 & 18.8 & 49.0 & 12.5 & 42.7 \\
\textbf{Ours} & \textbf{39.6} & \textbf{59.4} & \textbf{29.2} & \textbf{50.0} & \textbf{29.2} & \textbf{50.0} \\
\bottomrule
\end{tabular}
\vspace{-10pt}
\caption{\mt{Accuracy on the same 96 clips when all 97 EPIC-KITCHENS-100 verb classes are used as candidates, \ie, the protocol used for DEHOI.}}
\label{tab:human_vs_models_97}
\end{table}

\begin{table}[t]
\centering
\small
\setlength{\tabcolsep}{5pt}

\resizebox{\linewidth}{!}{
\begin{tabular}{l cc ccc ccc ccc ccc}
\toprule
\multirow{3}{*}{Method} 
& \multicolumn{2}{c}{EgoMCQ} 
& \multicolumn{6}{c}{EK-MIR} 
& \multicolumn{6}{c}{EK-MIR-Unseen} \\
\cmidrule(lr){2-3} 
\cmidrule(lr){4-9}
\cmidrule(lr){10-15}
& \multirow{2}{*}{Inter} & \multirow{2}{*}{Intra}
& \multicolumn{3}{c}{mAP} 
& \multicolumn{3}{c}{nDCG}
& \multicolumn{3}{c}{mAP} 
& \multicolumn{3}{c}{nDCG} \\
\cmidrule(lr){4-6} \cmidrule(lr){7-9}
\cmidrule(lr){10-12} \cmidrule(lr){13-15}
& & 
& V-T & T-V & Avg 
& V-T & T-V & Avg
& V-T & T-V & Avg 
& V-T & T-V & Avg \\
\midrule
LaViLa~\cite{zhao2023learning} 
& 94.2 & 63.2 
& 39.7 & 31.7 & 35.7 
& 36.1 & 33.2 & 34.6
& 33.8 & 41.6 & 37.7 
& 30.6 & 23.7 & 27.2 \\
Helping Hands~\cite{zhang2023helping}
& 94.5 & 63.0 
& 42.3 & 32.7 & 37.5 
& 39.3 & 36.2 & 37.8
& 33.4 & 41.5 & 37.4 
& 31.9 & 25.7 & 28.8 \\
\midrule
Ours
& 93.9 & 62.5 
& 40.4 & 32.6 & 36.5 
& 40.1 & 37.5 & 38.8
& 33.7 & 41.0 & 37.4 
& 33.2 & 27.2 & 30.2 \\
\bottomrule
\end{tabular}
}
\caption{Performance comparison on EgoMCQ, EK-MIR, and EK-MIR-Unseen.}
\label{tab:action_recognition}
\end{table}

\section{Additional Experiments}
We assess action recognition performance on \textbf{EgoMCQ}~\cite{lin2022egocentric} and \textbf{EPIC-KITCHENS-MIR} (EK-MIR)~\cite{damen2022rescaling} to verify that models trained to reason from partial observations achieve comparable performance on standard action recognition benchmarks.
We also report EK-MIR results on the split containing unseen manipulation-object combinations to highlight the improved generalization enabled by our cue-aware training strategy.
\mt{We further evaluate whether the improved object-centric understanding learned from egocentric human interactions generalizes to robot manipulation using the \textbf{DROID}~\cite{khazatsky2024droid} dataset, hypothesizing that object-centric dynamics enable action recognition even when the acting agent shifts from human hands to robot manipulators.}

\subsection{Datasets and Metrics}
\subsubsection{EgoMCQ~\cite{lin2022egocentric}} is a multiple-choice benchmark built on Ego4D, where given a text query for the action being performed, models select the most relevant video from five candidates.
It consists of approximately 39K questions, with two settings: inter-video, where candidates are sampled from different videos, and intra-video, where candidates are temporally adjacent clips from the same video, posing a more challenging scenario. Performance is evaluated using top-1 accuracy.

\subsubsection{EPIC-KITCHENS-MIR~\cite{damen2022rescaling}.} is a multi-instance text-video retrieval benchmark built on the EPIC-KITCHENS-100 dataset~\cite{damen2022rescaling}. It consists of 9,881 pairs of videos alongside their corresponding descriptions.
Models are evaluated on both video-to-text and text-to-video retrieval, with the final score averaged between the two tasks. Evaluation metrics include mean Average Precision (mAP) and normalized Discounted Cumulative Gain (nDCG).

\mt{\subsubsection{DROID~\cite{khazatsky2024droid}} is a robot manipulation dataset containing 76k demonstration trajectories with RGB observations from left, right, and wrist views. Each demonstration is annotated with a language instruction, such as “Put the marker in the bowl.” We filter demonstrations whose instructions contain a single verb and select the top 20 most frequent verbs, resulting in 47k demonstrations. The task is verb classification as in DEHOI and performance is evaluated using accuracy.}

\subsection{Results}
For action recognition (\cref{tab:action_recognition}), our training strategy, which mitigates overfitting to dominant observations in Ego4D, such as hand-centric cues, leads to improved performance in unseen settings, as reflected by the nDCG gains on EK-MIR-Unseen. However, compared to baselines that more strongly exploit Ego4D-specific biases, our model yields slightly lower accuracy by $\simeq0.5\%$ on EgoMCQ, which is also constructed from Ego4D videos.
On EK-MIR, we observe a slight decrease in mAP alongside an improvement in nDCG. While mAP evaluates retrieval performance under a binary relevance assumption, nDCG captures graded relevance by emphasizing whether semantically closer captions are ranked higher. This divergence suggests that, although the model retrieves fewer exact verb-noun matches at the top ranks, it more consistently prioritizes captions that are semantically closer to the query. Overall, these results indicate a shift from strict label-level matching toward finer-grained semantic ranking of hand-object interaction cues, aligning with the goal of learning more robust and less shortcut-driven representations.

\mt{\noindent\textbf{Robot manipulation recognition.}
\label{subsec:droid_supp}
\cref{tab:droid} shows the zero-shot performance of verb classification on robot manipulation videos from the DROID dataset~\cite{khazatsky2024droid}. Due to the substantial domain shift caused by differences in viewpoint and agent embodiment, all baselines show limited performance. Nevertheless, our model achieves the best performance across all viewpoints, suggesting that improved reasoning from object-centric cues, enabled by our training approach, leads to better generalization to manipulation videos performed by non-human agents.
}

\begin{table}[t]
\centering
\small
\setlength{\tabcolsep}{3pt}
\begin{tabular}{lcccccc}
\toprule
\multirow{2}{*}{Method}
& \multicolumn{2}{c}{Left View}
& \multicolumn{2}{c}{Right View}
& \multicolumn{2}{c}{Wrist View} \\
\cmidrule(lr){2-3} \cmidrule(lr){4-5} \cmidrule(lr){6-7}
& Top1 & Top5 & Top1 & Top5 & Top1 & Top5 \\
\midrule
LaViLa~\cite{zhao2023learning} & 2.9 & 20.1 & 2.9 & 20.3 & 2.6 & 17.3 \\
Helping Hands~\cite{zhang2023helping} & 4.0 & 25.8 & 4.3 & 25.6 & 3.2 & 18.9 \\
\midrule
\textbf{Ours} & \textbf{4.7} & \textbf{26.0} & \textbf{4.7} & \textbf{26.1} & \textbf{3.4} & \textbf{20.6} \\
\bottomrule
\end{tabular}
\vspace{-10pt}
\caption{\mt{Results on the DROID~\cite{khazatsky2024droid} robot manipulation benchmark.}}
\label{tab:droid}
\end{table}

\begin{figure}
    \centering
    \includegraphics[width=1\linewidth]{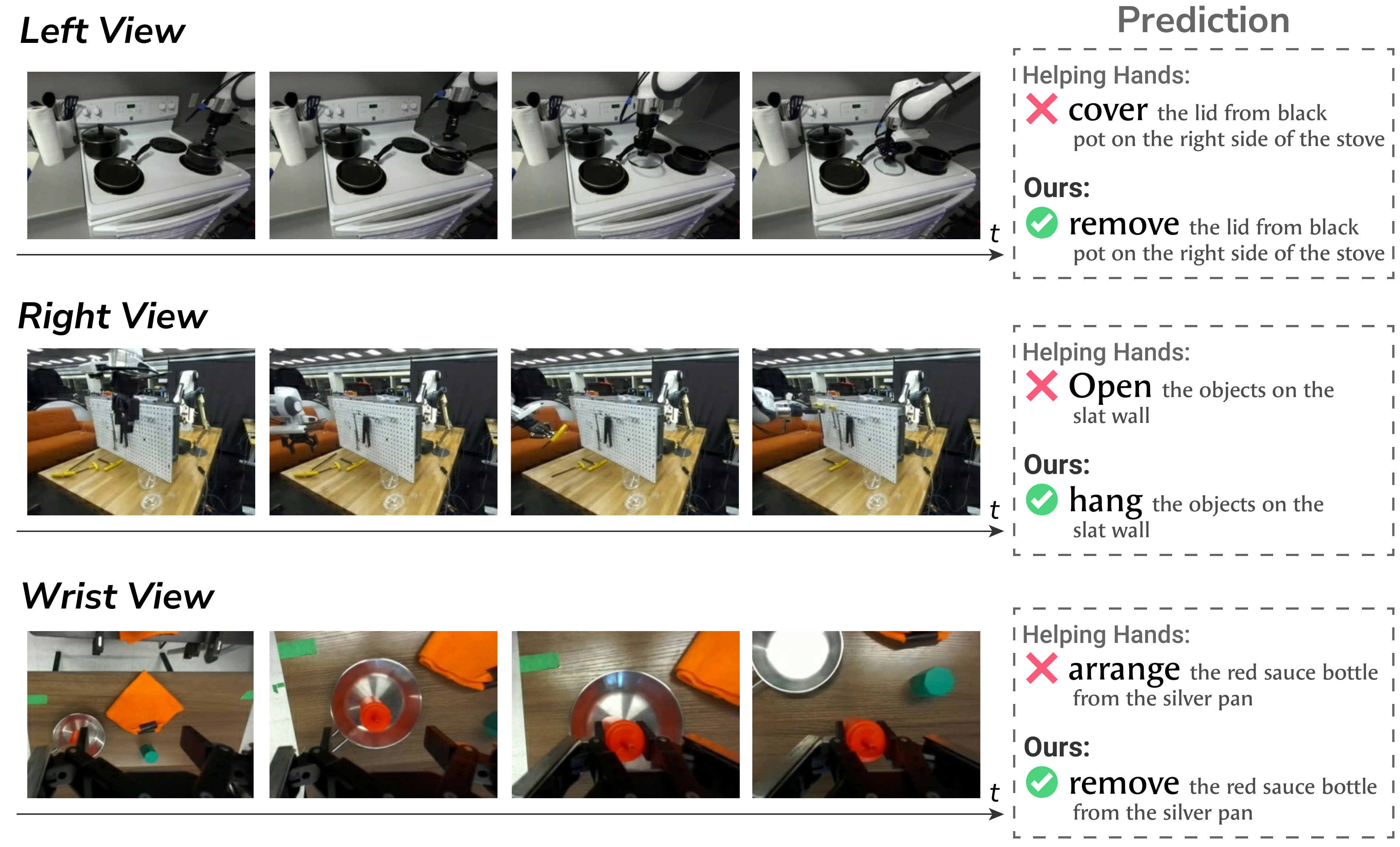}
    \caption{Qualitative results of action recognition from DROID~\cite{khazatsky2024droid} videos.}
    \label{fig:qualitative_droid}
\end{figure}

\mt{\cref{fig:qualitative_droid} further shows qualitative results on videos captured from the three viewpoints of DROID, \ie, left side, right side, and wrist view.
Our model correctly captures the underlying interaction dynamics, whereas existing models often fail and may even predict the opposite action (\eg, \emph{remove} vs.\ \emph{cover} in the top example).}

\end{document}